\documentclass[10pt,twocolumn,letterpaper]{article}

\usepackage[pagenumbers]{cvpr} %

\usepackage{microtype}

\definecolor{cvprblue}{rgb}{0.21,0.49,0.74}
\usepackage[pagebackref,breaklinks,colorlinks,allcolors=cvprblue]{hyperref}

\usepackage{algorithm}
\usepackage{algpseudocode}

\title{Procedural Adversarial Learning}

\author{Jack Nugent and Jia Deng\\
Department of Computer Science, Princeton University\\
{\tt\small \{jacknugent,jiadeng\}@princeton.edu}
}

\begin{document}
\maketitle

\begin{abstract}
For many computer vision tasks, neural networks can be trained to produce a low average error, yet they still fail on seemingly simple scenes. Reducing these failures requires first finding them, but they may not be in a fixed dataset. We study procedural adversarial learning (PAL), a framework of minimax training and adversarial evaluation which searches the input space of a procedural generator for difficult instances. The procedural generator is an interpretable program mapping a low-dimensional vector of control parameters (e.g.\@ layout, lighting, pose) to a rendered image and its label. We introduce the Simple-PAT algorithm, which uses zeroth-order optimization in the procedural generator's parameter space to propose adversarial training samples that are generated on demand. We also create PAL-Dojo, a collection of efficient procedural generators for surface normal estimation, image segmentation, and human pose estimation. 
Against our strongest black-box adversary with 150M aggregate queries, Simple-PAT reduces worst-case error by a geometric mean of $2.11\times$ compared to standard training and $1.75\times$ compared to the strongest robust training baseline. Against adversaries of varied strengths Simple-PAT is more robust than all baselines.
Code is available at \url{https://github.com/princeton-vl/procedural-adversarial-learning}.
\end{abstract}

\section{Introduction}
\label{sec:intro}

Vision models deployed in the wild are often relied on to produce high-quality predictions in all situations. This is particularly important for self-driving cars, surgical robots, and heavy machinery, where failures in perception can lead to dangerous and costly mistakes. However, even well-trained neural networks often fail on simple scenes or natural adversarial examples \citep{hendrycks2021natural,alcorn2019strike}. This is not easily solved by scaling data or model size, as foundation VLMs also struggle on examples that are easy for humans \citep{naturalbench}. These failures motivate training and evaluation techniques that explicitly account for worst-case error.

How can we train a model that has low worst-case error? A natural and established idea is to define a minimax learning objective: minimizing the worst-case error, defined as the maximum error over all possible instances of the task space. While simple to state, this minimax objective is challenging to operationalize: typically, we are given a fixed finite sample of the task instances such as a set of labeled images. This sample does not allow us to reliably estimate the maximum error over all possible images since failure cases may be unseen. If we cannot reliably measure the worst case error, we cannot reliably improve it.  

To discover missed failure modes and reduce worst-case error, prior work in robust and adversarial machine learning has attempted to perturb existing images (e.g. \cite{goodfellow2014explaining,pmlr-v97-engstrom19a,hosseini2018semanticadversarialexamples}). However, such perturbations are usually limited to simple transformations such as color jitter and spatial cropping, and cover only a tiny subset of valid image transformations. 
This is because automatically generating physically valid perturbations of existing images that span viewpoint changes, dynamic lighting, and shape deformation remains an active research challenge plagued by its own failure modes. On the other hand, collecting additional real-world data can be slow and expensive. 

Thus, a key challenge of improving worst-case error lies in scalable discovery and generation of failure cases. 
In this work, we address this with Procedural Adversarial Learning (PAL), a framework consisting of procedural adversarial training (PAT) and procedural adversarial evaluation (PAE); the key idea is to adversarially select samples using a procedural generator.
During training, a learner updates the parameters of the network while an adversary searches for the worst case. For evaluation, a strong adversary searches the procedural parameter space for the task configuration that is most challenging for the model.

\begin{figure*}[ht]
    \centering
    \includegraphics[width=0.8\linewidth]{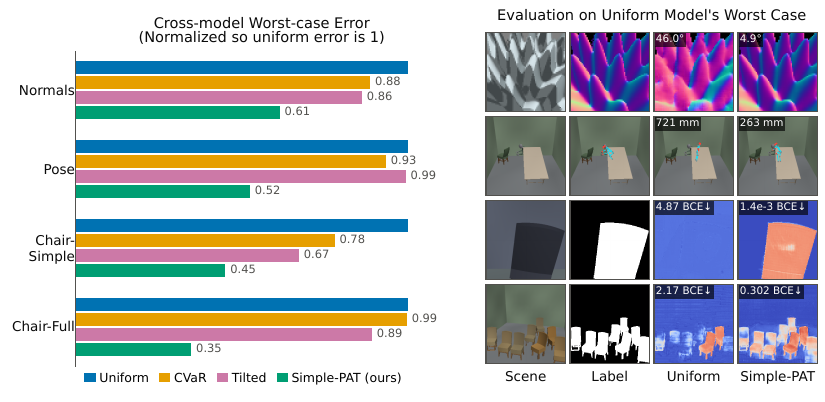}   
    \vspace{-1em}
    \caption{We compare Simple-PAT to training methods that uniformly sample data for each minibatch. Left: Simple-PAT reduces worst-case error more effectively than uniform training and alternative methods. Right: Simple-PAT models perform much better on cases uniformly trained models found challenging. Uniform refers to the ``Uniform (Muon)'' setting, which is the stronger of the uniform training methods.}
    \label{fig:fig1}
\end{figure*}

A procedural generator generates data through interpretable code: an image of a tree can be procedurally generated from scratch by code that simulates tree growth, bark material, light transport, and perspective projection. Procedural generation can create diverse photorealistic images~\citep{raistrick2023infinite,raistrick2024infinigen}, visual ground truth, and useful training data~\citep{ma2026simpleprocfullyproceduralsynthetic,yan2026makesgoodsynthetictraining}. 
Importantly, procedural generation enables scalable discovery and generation of failure cases. A well-designed procedural generator can offer a combination of properties unavailable from image-space perturbation: (1) wide coverage of task instances, (2) physically valid perturbations, (3) a small number of control parameters that help traverse the task space, (4) on-demand generation of new task instances. 

These properties make procedural generators an especially strong tool for PAL, since we can effectively search a visually complex collection of scenes. This distinguishes PAL from conventional adversarial learning, which searches a limited set of image perturbations. The evaluation also distinguishes PAL from Adversarial Domain Randomization \citep{khirodkar2021adversarialdomainrandomization}, which trains with the same minimax formulation over a simulator’s parameters but evaluates mean performance in a target domain.
In PAL, training seeks to minimize the worst case under the generator and evaluation searches for it directly. The challenge lies in developing training algorithms that reliably improve the worst case. This can be particularly difficult when failures occupy only a small region of the parameter space, or when learning one failure causes the model to forget scenes it previously handled.

We propose Simple-PAT, a simple algorithm for procedural adversarial training using adversaries that attack via zeroth-order optimization. Simple-PAT consists of a learner that trains the model through minibatch gradient descent and an adversary that uses Covariance Matrix Adaptation Evolution Strategy (CMA-ES) \cite{cmaes} to search on procedural controls to generate the next minibatch based on existing cases the model finds challenging.

To analyze PAT algorithms, we develop PAL-Dojo, a suite of procedural generators for three vision tasks: surface normal estimation, human pose estimation, and image segmentation. We design these generators to control the scenes’ layout, lighting, shapes, and poses using between 13 and 69 parameters. We prioritize simplicity and efficiency of generation to facilitate algorithmic development. 

We evaluate trained models on PAL-Dojo by treating them as a black box and adversarially attacking with a large query budget. We use a single-model adversary with 6M queries and a cross-model adversary with a total of 150M queries. In Figure \ref{fig:fig1}, we compare Simple-PAT to a naive learner that samples every batch uniformly from the parameter space (uniform training), and to two methods that reweight uniform samples based on their loss (CVaR and tilted ERM). Under the cross-model adversary, Simple-PAT lowers the worst case on every task, by a geometric mean of $2.11\times$ relative to uniform training and $1.75\times$ relative to the strongest other method on each task. We also compare methods across adversaries of different strengths, measuring strength against a standard baseline. As the attack strengthens, Simple-PAT's worst case grows more slowly than every baseline on every task. Finally, we evaluate the robustness of SOTA foundation models using the PAL-Dojo generators, finding severe failure cases with our adversary. This demonstrates that PAL-Dojo, though visually simple, provides sufficient challenge to SOTA models.

In summary, our contributions are threefold: (1) Procedural Adversarial Learning (PAL), a framework that pairs minimax training under a procedural generator with worst-case evaluation by adaptive adversaries of varied strengths; (2) Simple-PAT, a simple yet effective procedural adversarial training algorithm; (3) PAL-Dojo, a suite of simple, efficient but sufficiently hard procedural generators for vision tasks, designed to facilitate the study of PAL.

\section{Related Work}

\noindent \textbf{Natural Adversarial Examples.} Image classifiers and vision-language models are known to fail on unmodified natural images that humans label easily \citep{hendrycks2021natural,naturalbench}. Search can also reveal these failures systematically, by adjusting the pose \cite{alcorn2019strike}, or other elements including object placement and lighting \citep{zeng2019adversarial,leclerc20223db}.

\noindent \textbf{Adversarial training and robust objectives.} Classical adversarial examples were defined as small pixel perturbations \citep{szegedy2014intriguingpropertiesneuralnetworks,goodfellow2014explaining}, and adversarial training addresses this by minimizing loss under perturbations within an $L^\infty$ ball around each image \citep{madry2017towards}. However, these perturbations are limited and the inner optimization loop during training can be costly. Alternative methods instead emphasize difficult examples through a distributionally robust training objective. CVaR averages the worst $\alpha$ fraction of samples \citep{rockafellar2000optimization,levy2020large}, and tilted ERM weights each example in proportion to $\exp(\beta \ell)$ for a loss $\ell$ and hyperparameter $\beta$ \citep{tiltederm}.

\noindent \textbf{Adversaries in Reinforcement Learning and Control.} Research in reinforcement learning has also searched for challenging training instances.
PAIRED trains a level generator to maximize the agent's regret \citep{dennis2020emergent}. ACCEL applies random mutations to high-regret levels to generate new difficult instances \citep{parker2022evolving}. Importantly, both optimize regret instead of loss since directly optimizing loss would produce impossible tasks in their environments. Neither work attempts to evaluate the worst-case performance over the entire parameter space. 

Adversarial game formulations have also been used to develop safe and robust learning-based controllers, through minimax policy training against disturbance forces \citep{pinto2017robustadversarialreinforcementlearning} and safety control against worst-case disturbances \citep{fisac2018generalsafetyframeworklearningbased}.

\noindent\textbf{Adversarial Domain Randomization.} Adversarial Domain Randomization (ADR) \citep{khirodkar2021adversarialdomainrandomization} trains a learner against a simulator in a minimax game, with a policy over quantized simulation parameters rewarded by the learner's loss. In ADR's experiments, this policy controls a few coarse parameters (e.g. object class and placement), while lighting, texture, and pose are randomly sampled. 
ADR demonstrates that such a policy identifies distributions that produce useful samples, such as occluded objects and difficult placements, and outperforms domain randomization \citep{tobin2017domainrandomizationtransferringdeep} on mean performance in the target domain. In contrast, PAL's evaluation searches for the worst case in the same parameter space used for training.

\section{Procedurally Generated Tasks}
\label{sec:procedural_generators}
We define a procedural generator as a deterministic map $g$ from a parameter vector $u \in [0,1]^d$ to a rendered image $x_u$ and its ground-truth label $y_u$. We can view this as an infinite source of controllable data for both training and evaluation. In contrast, a fixed dataset can be viewed as $N$ queries of the generator $g$ determined by some unknown selection process. Access to the generator has the following consequences:
\begin{enumerate}
    \item The training data need not be fixed in advance. At each training step we select $B$ parameter vectors to form the next batch, and this selection can depend on the current model. A fixed dataset only allows choices among its $N$ members, and must repeat after a number of training steps. If certain regions of the parameter space are particularly valuable, the number of unique useful samples may be quite small. Given access to $g$, adaptive training can concentrate on these useful regions of the parameter space and present the model with new examples at every step.
    \item The worst case can be searched for over the entire parameter space. We consider the absolute worst-case
    \[
        W(\theta) = \max_{u \in [0,1]^d} \ell(\theta, u) \qquad
        \ell(\theta, u) = L\big(f_\theta(x_u), y_u \big)
    .\]
    For training and evaluation we can approximate this by sampling. However, the structure of the procedural generator enables more effective search. Each parameter controls a perceptual quality of the scene (e.g. lighting, object shape, layout). Since these qualities also relate to the difficulty of the scene, we expect the difficulty to vary reasonably smoothly in the parameter space. This enables search algorithms to identify the challenging cases much more quickly than random sampling.

\end{enumerate}

\begin{figure*}[t]
    \centering
    \includegraphics[width=\linewidth]{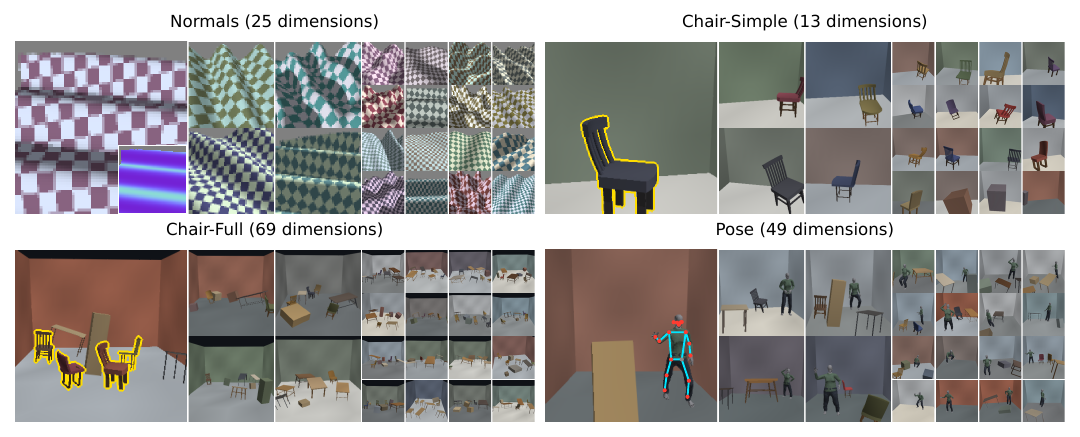}
    \caption{Samples from our PAL-Dojo procedural generators. The leftmost panel of each task shows a visualization of the ground truth labels.}
    \label{fig:env_samples}
\end{figure*}

\noindent
In order to make $W(\theta)$ meaningful, it must be the case that for every $u \in [0,1]^d$, $y_u$ is solvable given $x_u$. Otherwise the worst case will be an ambiguous scene. For PAL-Dojo, we build four procedural generators to represent a variety of vision tasks (samples shown in Figure \ref{fig:env_samples}). We design them to be solvable and extremely efficient so that we can generate data before each training step and practically search for the worst cases. 

\subsection{Normal Estimation}

For the normal estimation task, the model must estimate the 3-dimensional surface normal vector at every pixel on a curved surface. This surface is a heightfield determined by a mixture of sinusoidal waves. The material is a checkerboard pattern rendered with Lambertian shading under a directional light with an ambient term. The presence of a repetitive checkerboard pattern makes normal estimation tractable, since any warping in the pattern is due to surface normal variation. The network must also learn to appropriately handle shadows and lighting cues.

The generator has 25 parameters, with 13 parameters specifying the waves, 2 describing the light azimuth and elevation, and 10 describing the checkerboard texture. This includes 24 continuous parameters, and 1 discrete parameter specifying the number of waves.

We use the metric of average angular error (in degrees) to describe models' performance.

\subsection{Object Segmentation}

Our object segmentation task asks models to classify each pixel as belonging to a chair object or not. The generator places chairs and distractor objects within a room while ensuring the chair is not occluded excessively. We have a Chair-Simple variant with 13 dimensions and a Chair-Full variant with 69 dimensions.

The procedural objects that can be placed in the room include chairs, tables, desks, and cubes. These objects are adapted from ProcFunc implementations of objects in Infinigen, with a custom backend designed for rapid generation \citep{procfunc}.
To keep the dimensionality manageable, we map three parameters to the inputs of each object's generator: one selects a discrete style, and two set all of its continuous inputs. We design this mapping so that uniform sampling reproduces Infinigen's distribution of each input (i.e., the marginal distribution of each Infinigen parameter). However, since there are only three parameters, we cover fewer combinations of inputs than Infinigen's independent sampling.

In the Chair-Simple task, we have 5 parameters for the room shape and camera placement, and we have 8 parameters for the object type, shape, and position.

For the Chair-Full task, we place up to 8 objects on a $16\times16$ grid of positions on the floor. Each
 object has a parameter $d_i$ that selects a target cell, with cells ordered so that consecutive cells are
 adjacent on the floor. Objects are placed sequentially. For object $i$, we identify the valid grid cells given the objects already placed, and place the object in the valid cell nearest to its target cell.
We determine cell validity based on criteria of non-intersection, the boundaries of the room, and non-occlusion. We restrict occlusion to avoid the case where an object is visible but its class is not identifiable (e.g. a chair where you can only see its legs).
This arrangement procedure ensures that the problem is tractable, but also introduces discontinuities and sensitive parameters. \looseness=-1

We evaluate models based on average pixel binary cross entropy (BCE) loss. 

\subsection{Human Pose Estimation}
The human pose estimation task builds on the scene generator from the object segmentation task. It includes three object slots, and one human slot. 

The human figure uses the Anny model with clothing from MakeHuman \citep{anny, makehuman}. We include five identity coordinates (gender, age, muscle, weight, proportions), and 9 pose coordinates.
For pose, we perform archetypal analysis dimensionality reduction on the poses from the CMU mocap database \citep{cmumocap}, and we remove two archetypes with upside-down poses. The 9 pose coordinates determine the mixing weights on the remaining 10 archetypes. 
Our goal is to represent plausible poses instead of every possible joint configuration, which would produce self-intersection and other non-physical configurations.

Similar to the object segmentation task, we ensure the human figure is visible and unoccluded by only placing it in valid cells.

For pose estimation, we use root-aligned mean per-joint position error (MPJPE) as our metric:
$
    MPJPE = \frac{1}{|V|} \sum_{j \in V}\| (\hat{p}_j - \hat{p}_{root}) - (p_j - p_{root})\|_2
$
where $p_{root}$ is the average of the left and right hip positions. We only calculate the error for joints visible in the scene.

\subsection{Performance Characteristics}

\begin{table}[h]
    \centering
    \small
    \begin{tabular}{cccc}
         Task & Dimension & Resolution & Scenes/second \\
         \hline
         Normals & 25 & $64^2$ & 8700 \\
         Chair-Simple & 13 & $256^2$ & 870 \\
         Chair-Full & 69 & $256^2$ & 350 \\
         Pose & 49 & $256^2$ & 320 \\
         
    \end{tabular}
    \caption{Performance of procedural generators. Mean over 100 batches measured on an NVIDIA RTX4090 GPU at a batch size of 256. The chair and pose generators build scenes using 6 CPU worker processes.}
    \label{tab:perf}
\end{table}

In Table \ref{tab:perf} we show the performance of our procedural generators. Scene generation and rendering are optimized for batched throughput and GPU efficiency, making them ideal for neural network training pipelines.

\section{Adversarial Training}

We design our adversarial training procedure Simple-PAT around three principles: (1) \textit{coverage:} the model must be presented with examples near the worst-case example (2) \textit{diversity:} each batch the model receives should contain maximally distinct examples (3) \textit{computation reuse:} samples which we render and evaluate should be used to update the network. Though simple, these principles allow us to design an algorithm whose training throughput is comparable to uniform training, while achieving a large reduction in evaluation worst-case error.

Algorithm \ref{alg:adv} describes Simple-PAT. We maintain $k$ CMA-ES instances $\pi_1,\dots,\pi_k$. At each step we select a batch using $n_0$ uniform samples and $\lambda$ samples from each $\pi_i$. We run a forward pass on every data point and use the results to update the CMA-ES parameters. We then run a backward pass and optimization step, as in standard neural network training. Every $R$ steps we reset the weakest CMA-ES instances as determined by a fitness function.

\begin{algorithm}[H]
\caption{Simple-PAT}
\label{alg:adv}
\small

\begin{algorithmic}[1]
\Require generator $g:[0,1]^d\!\to\!\mathcal{X}\times\mathcal{Y}$, model $f_\theta$, per-scene loss $\ell$, reset period $R$, sampler parameters $n_0, \lambda$
\State init $k$ CMA-ES distributions $\pi_1,\dots,\pi_k$ on $[0,1]^d$
\For{$t = 1,\dots,T$}
\State $U \gets \{u \sim \pi_i\}_{i\le k}^{\times\lambda} \cup \{u \sim \mathcal{U}[0,1]^d\}^{\times n_0}$
\State $(x, y) \gets g(U)$
\State $\hat{y} \gets f_\theta(x)$ \Comment{forward, per scene}
\State $f \gets \Call{Fitness}{\hat{y}, y}$
\State $\pi_i \leftarrow \Call{UpdateCMAES}{\pi_i, f}$ for all $i$
\State $\nabla \gets \nabla_\theta\, \tfrac{1}{B}\textstyle\sum \ell(\hat{y},y)$ \Comment{backward}
\State $\theta \gets \mathrm{OptimStep}(\theta, \nabla)$
\If{$t \bmod R = 0$}
\State $\{\pi_i\} \leftarrow \Call{ResetWeakest}{\{\pi_i\}}$
\EndIf
\EndFor
\end{algorithmic}
\end{algorithm}

\noindent\textbf{CMA-ES Optimization.} CMA-ES is a zeroth-order optimizer that uses a multivariate normal distribution to sample candidate solutions to the optimization problem \citep{cmaes}. This is amenable to our setup due to batched computation and diversity in samples. At each step, $\lambda$ points are sampled from $N(\mu, \Sigma)$, these points are evaluated by fitness, and $\mu$ and $\Sigma$ are updated. We constrain $\Sigma$ to be diagonal since $\lambda$ is relatively small. For fitness, we use the evaluation metric of each task. This may differ from the training loss function due to stability or smoothness reasons.

We also change standard CMA-ES by pooling updates over 2 generation steps (i.e. $\mu$ and $\Sigma$ are updated every other training step). This increases the number of samples used to compute the updates and captures some of the non-stationarity in the objective. If worst case is moving in parameter space due to neural network updates, this pooling operation discourages CMA-ES from over-concentrating at an optimum which exists only transiently. 

\noindent\textbf{CMA-ES Resets.} We find it is helpful to reset the CMA-ES instances periodically. Individual instances may be stuck in a local maximum, or multiple instances could converge to the same point and reduce the batch's diversity. For each $\pi_i$ we define a fitness function based on loss and separation, and we replace the least fit fraction of $\pi_i$ instances every $R$ training steps. This should leave searchers that find high loss and are far apart from others.

\section{Adversarial Evaluation}
To evaluate the robustness of training methods, we search the procedural parameter space for the worst case. We treat the trained network as a black box and optimize for $\max_u \ell (\theta, u)$ using standard techniques. This directly assesses the efficacy of the training method at achieving its minimax objective, and is enabled by the fact that the procedural generator produces accurate ground truth for every input parameter vector.

\subsection{Theoretical Properties}
One benefit of a procedural generator is that by sampling enough scenes we can approximate the true worst case. This holds under mild assumptions that fit our procedural generators. In particular, any optimizer whose queries eventually cover the parameter space converges to the worst case; this includes uniform sampling, DIRECT-L, and TuRBO with restarts. See Appendix~\ref{sec:app_convergence} for additional details.

Unfortunately, the expected number of samples to approach the worst case may be arbitrarily large. We are unable to make any assumptions about Lipschitz continuity or other properties that would give convergence rates. Therefore, we cannot determine whether an optimizer has converged.

\subsection{Practical Considerations}
\label{sec:eval_practical}
The above properties present some practical challenges. We would ideally like to measure the true worst-case performance of models, but cannot always do so with finite compute. Conventional adversarial evaluation using $L^p$ perturbations faces the same challenges; we adopt similar principles to \citet{carlini2019evaluatingadversarialrobustness}. We establish an evaluation procedure applied identically to each model, and define the \textit{evaluation worst case} to be the worst case found by this procedure. Our claims about training efficacy relate only to this evaluation worst case.

The purpose of evaluation is to make comparisons between methods. To account for randomness, we train and evaluate multiple seeds for each method and use a two-sided Mann-Whitney U test for comparisons. This test makes no assumptions about the shape of the distribution of the evaluation worst cases. A significant result means that one method consistently ranks better than another under the same evaluation procedure.

To reduce bias in comparisons, we select a variety of attack strategies and use an evaluation budget larger than our training budget. We include optimizers with distinct characteristics, we perform a cross-model evaluation, and we include uniform random sampling baselines. We also plot optimizer convergence curves for each trial. None of these ensure that we find the worst case, but our goal is for the search to be equally effective against every training method.

Finally, given a fixed setup of a network, procedural generator, and training budget, it might not be reasonable for models to be resistant to all possible attacks. In addition to comparing efficacy against the strongest attacks, we can compare each model's efficacy against attacks of a particular strength. We consider each pair of an optimizer and a search budget as an attack, and define an attack's strength as the worst-case error it finds against a baseline (uniform Muon in our case). We can then compare how quickly each training method's worst case grows as the attack becomes stronger. 

\section{Experiments}

\subsection{Setup}
On the PAL-Dojo tasks, we compare Simple-PAT adaptive training to training techniques that randomly select samples. Please see Appendix~\ref{sec:app_hparam} for additional details.

\noindent\textbf{Baselines.}
\begin{itemize}
\item 
\textit{Uniform training}, which samples each batch of $B$ parameter vectors uniformly from $[0,1]^d$. This is equivalent to training on an infinite fixed dataset with every
 point weighted equally. We train uniform models with both Muon and Adam. 
\item
\textit{CVaR}~\citep{rockafellar2000optimization,levy2020large}, which draws a uniform batch but backpropagates only the worst $\alpha$ fraction of its scenes. We use $\alpha=0.05$, so 13 of 256 scenes contribute to each update. 
\item
\textit{Tilted ERM}~\citep{tiltederm}, which draws a uniform batch and weights each scene's gradient by $\mathrm{softmax}(\beta\ell)$ over the batch. As $\beta\to\infty$, this approaches the gradient of the batch's worst case. We choose $\beta$ for each task so that the effective sample size of the weights matches that of CVaR.
\end{itemize}

\noindent\textbf{Evaluation search.} We search for each model's worst case using six black-box optimizers: CMA-ES~\citep{cmaes}, DIRECT-L~\citep{directL}, TuRBO~\citep{turbo}, differential evolution~\citep{storn1997differential}, a genetic algorithm~\citep{pymoo}, and uniform random sampling. Each uses $10^6$ queries per model. The single-model worst case is the largest error found by any of these searches. The cross-model worst case evaluates each model on the worst cases found against every other model on the same task. Across 25 models, there are $1.5 \times 10^8$ pooled queries for each task. Though we use CMA-ES for training and as an evaluation procedure, the evaluation search is much stronger -- it uses a population size of 512 instead of 28, a full covariance estimate instead of diagonal, IPOP restarts, and it optimizes against a fixed network instead of one that changes at each step.

\noindent\textbf{Hyperparameters.} Every method trains networks of 5M--7.8M parameters with Muon~\citep{jordan2024muon} (or Adam~\citep{adam} where stated) for 20,000 steps at a batch size of 256, a total of $5.12\times10^6$ training samples. In each Simple-PAT batch, $n_0=32$ scenes are sampled uniformly and 224 come from $k=16$ CMA-ES instances with $\lambda=14$ samples each.
 
\subsection{Robustness to the Strongest Adversary}
\label{sec:cross_model_eval}
\begin{table*}[h]
\centering
\small
\setlength{\aboverulesep}{0pt}\setlength{\belowrulesep}{0pt}
\renewcommand{\arraystretch}{1.18}
\setlength{\tabcolsep}{3.5pt}
\begin{tabular}{@{}l | r@{\,}l r | r@{\,}l r | r@{\,}l r | r@{\,}l r@{}}
\toprule
 & \multicolumn{3}{c|}{Normals, angle ($^\circ$)} & \multicolumn{3}{c|}{Chair-Simple, BCE} & \multicolumn{3}{c|}{Chair-Full, BCE} & \multicolumn{3}{c}{Pose, MPJPE (mm)} \\
\cmidrule(lr){2-4}\cmidrule(lr){5-7}\cmidrule(lr){8-10}\cmidrule(lr){11-13}
Method & \multicolumn{2}{c}{Worst $\downarrow$} & Median $\downarrow$ & \multicolumn{2}{c}{Worst $\downarrow$} & Median $\downarrow$ & \multicolumn{2}{c}{Worst $\downarrow$} & Median $\downarrow$ & \multicolumn{2}{c}{Worst $\downarrow$} & Median $\downarrow$ \\
\midrule
Uniform (Adam) & 50.55 & {\scriptsize$\pm$0.70}$^*$ & 1.6 & 4.66 & {\scriptsize$\pm$0.15}$^*$ & 2.1e-5 & 3.22 & {\scriptsize$\pm$0.17}$^*$ & 7.5e-5 & 927 & {\scriptsize$\pm$46}$^*$ & 26 \\
Uniform (Muon) & 43.4 & {\scriptsize$\pm$3.0} & \textbf{0.89} & 4.778 & {\scriptsize$\pm$0.056}$^*$ & \textbf{5.5e-6} & 2.50 & {\scriptsize$\pm$0.27}$^*$ & \textbf{1.9e-5} & 742 & {\scriptsize$\pm$33}$^*$ & \textbf{16} \\
Tilted & 37.4 & {\scriptsize$\pm$1.5} & 3.7 & 3.21 & {\scriptsize$\pm$0.13}$^*$ & 3.5e-4 & 2.23 & {\scriptsize$\pm$0.14}$^*$ & 1.7e-4 & 740 & {\scriptsize$\pm$100}$^*$ & 38 \\
CVaR & 38.4 & {\scriptsize$\pm$2.2} & 1.8 & 3.72 & {\scriptsize$\pm$0.17}$^*$ & 8.7e-5 & 2.49 & {\scriptsize$\pm$0.41}$^*$ & 5.1e-5 & 691 & {\scriptsize$\pm$60}$^*$ & 40 \\
\textbf{Simple-PAT} & \textbf{27} & {\scriptsize$\pm$11} & 1.7 & \textbf{2.16} & {\scriptsize$\pm$0.26} & 1.9e-5 & \textbf{0.87} & {\scriptsize$\pm$0.43} & 5.6e-5 & \textbf{389} & {\scriptsize$\pm$46} & 27 \\
\bottomrule
\end{tabular}

\caption{Cross-model evaluation worst case for various training methods. Each entry is the mean of 5 seeds
with $\pm$ 1 standard deviation shown for the worst cases. All entries marked with ($^*$) indicate that Simple-PAT is significantly better at $p<0.05$ on the two-sided Mann-Whitney U test.
}
\vspace{-0.5em}
\label{tab:main}
\end{table*}

Table \ref{tab:main} shows robustness against the cross-model adversary, which is the strongest of our adversaries (see Appendix~\ref{sec:app_single} for single-model adversary results). Simple-PAT has the lowest mean worst case on all four tasks. Compared to uniform Muon, it reduces
 the cross-model worst case error by $1.63\times$ on normals, $2.22\times$ on chair-simple, $2.87\times$ on chair-full,
 and $1.91\times$ on pose. Against the strongest baseline for each task, the reductions are $1.40\times$,
 $1.49\times$, $2.55\times$, and $1.78\times$.

Under the Mann-Whitney U test, thirteen of the sixteen comparisons are significant at $p < 0.05$, each with $p=0.0079$. In these cases, all five Simple-PAT seeds have a lower worst case than all five seeds of the baseline. On normals, Simple-PAT has the lowest mean worst case, but its comparisons with Muon, tilted, and CVaR are not significant ($p = 0.056, 0.15, 0.15$). Simple-PAT has one of the five seeds with a worst case of $44.3^\circ$, while the remaining seeds all have worst cases between $19.1^\circ$ and $28.3^\circ$. Since this single seed overlaps with the worst cases of other methods, our $n=5$ comparisons are not significant.

\noindent\textbf{Typical case cost.} Our primary goal is to improve worst-case performance, but Simple-PAT generally performs well on typical cases as well. On Normals, Chair-Simple, and Pose, Simple-PAT has the best median performance among the robust training methods. Moreover, the magnitude of the median errors is quite small. On the pose task, Simple-PAT is 1.1cm worse than uniform Muon in the median case while improving the evaluation worst case by 35cm. On Chair-Simple and Chair-Full, Simple-PAT's median error is comparable to being uncertain about a few pixels (e.g. a model that predicts probability of 0.5 on six pixels in a $256^2$ image has a BCE of at least $6.3 \times 10^{-5}$). 

\subsection{Robustness Across Adversary Strengths}

\begin{figure}
    \centering
    \includegraphics{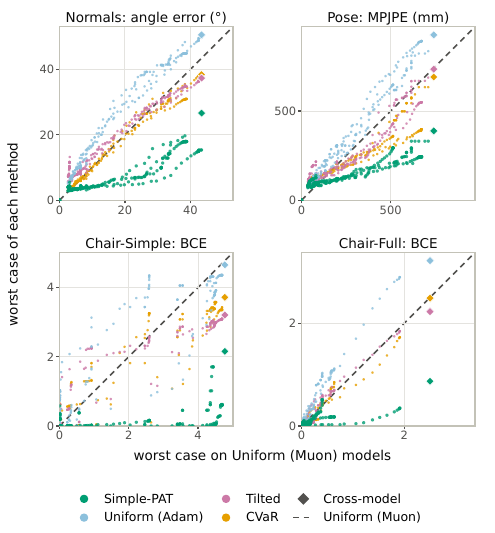}
    \caption{Worst case against each training method, with strength measured as the worst case against uniform Muon. Each dot represents an $(\text{optimizer}, \text{budget})$ pair and shows the mean worst case over all five seeds.}
    \label{fig:adversary_strength}
\end{figure}

Figure \ref{fig:adversary_strength} compares the robustness of all training techniques to adversaries of different strengths. At low adversary strengths on Chair-Full, Pose, and Normals, the training methods are largely indistinguishable. At high adversary strengths, we see that Simple-PAT is more robust than alternate methods. For Chair-Simple, the distinction between Simple-PAT and alternate methods is apparent even at low adversary strengths. However, when the Chair-Simple worst case on uniform Muon exceeds 4, we see that Simple-PAT's worst case increases rapidly. In this sense, its defense is ``broken'' by these attacks (TuRBO and GA), though its final error is still lower than other methods.

To analyze each method's overall trend, we fit each seed's worst case against adversary strength by least squares through the origin. By defininition uniform-Muon has a slope of 1. Averaging across seeds the slopes on the four tasks range from 0.08 to 0.45 for Simple-PAT, 0.70 to 1.02 for tilted and CVaR, and 0.94 to 1.65 for uniform Adam. Thus, robust reweighting methods provide only modest robustness improvements over uniform Muon as the attack strengthens. In contrast, Simple-PAT degrades substantially less against stronger attacks. Appendix \ref{sec:app_robustness_strength} shows the robustness to each kind of attack.

\subsection{Simple-PAT Instability}
Since the original comparison between Simple-PAT and baselines on the Normals environment was not significant, we ran a fifteen seed replication to better compare performance. Our findings are (1) Simple-PAT exhibits training instabilities, (2) with $n=15$ seeds, Simple-PAT is better than baselines at $p<0.01$, and (3) with the measured variance, our original five seed test had statistical power of only 0.3\footnote{See Appendix~\ref{sec:app_stats}.}, suggesting more seeds were needed.

\begin{table}
\centering\small
\begin{tabular}{@{}l c r@{\,}l c c@{}}
\toprule
Study & $n$ & \multicolumn{2}{c}{Worst ($^\circ$)} & Max $p$-value & Seeds $\geq 40^\circ$ \\
\midrule
Original    & 5  & 27 & {\scriptsize$\pm$11} & 0.15   & 1/5  \\
Replication & 15 & 27 & {\scriptsize$\pm$12} & 0.0020 & 2/15 \\
\bottomrule
\end{tabular}
\caption{Simple-PAT on Normals in the original five-seed study and the fifteen-seed replication. Worst: mean $\pm$ 1 standard
deviation over seeds. Max $p$: largest two-sided Mann-Whitney $p$ over the four baselines.}
\label{tab:replication}
\vspace{-1em}
\end{table}

\begin{figure*}
    \centering
    \includegraphics[width=0.9\linewidth]{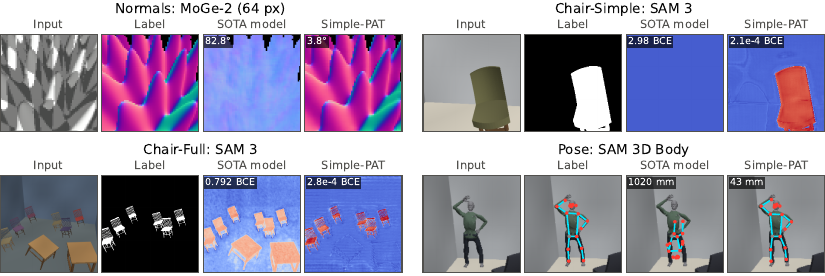}
    \caption{Failure cases of SOTA foundation models. We show the worst case found from a CMA-ES search against the foundation models with budget $10^6$ ($10^5$ for SAM 3D Body).}
    \label{fig:sota_worst}
\end{figure*}

As seen in Table \ref{tab:replication}, Simple-PAT's mean worst case and the standard deviation between seeds is similar at $n=5$ and $n=15$. Despite a median worst case of $24.5^\circ$ between seeds, there are 2/15 seeds with a worst-case error above $40^\circ$. When we evaluate these $>40^\circ$ scenes on other Simple-PAT seeds, their mean error is only $15.7^\circ$. This suggests that some Simple-PAT training can fail on difficult cases unique to a particular run. Future work can address this by improving the coverage of the generator's parameter space. In our $n=15$ study, we do find that Simple-PAT is significantly better than baselines with $p<0.01$ for all comparisons. We find ratios of 1.62 compared to uniform Muon and 1.35 compared to the next-best baseline; the original study found ratios 1.63 and 1.40.

\subsection{Worst Case Evaluation of State-of-the-Art Models}

We also use our evaluation framework to understand the robustness of SOTA foundation models. 
Though these models have not been trained on our data, the images are perceptible to humans and we would expect them to generalize here. Moreover, foundation models benefit from diverse training data and may have learned rules for these tasks that generalize to all settings. We evaluate MoGe-2~\citep{wang2026moge} for normal estimation, SAM 3~\citep{carion2026sam} for segmentation, and SAM 3D Body~\citep{yang2026sam} for pose estimation.

For MoGe-2 and SAM 3, we use the predicted normals and class probabilities of the model to compute our loss. We give SAM 3 the prompt ``chair'' and clamp each pixel's predicted probability to $[10^{-6}, 1-10^{-6}]$, so that the loss stays finite.
For SAM 3D Body, in addition to aligning the root as in MPJPE, we also align the scale of the prediction. This is necessary since the scale of these synthetic scenes is ambiguous without being trained on our specific generator.

In Figure \ref{fig:sota_worst} we visualize the worst cases found for the SOTA foundation models on our tasks. For the normals and chair-simple tasks, the search identifies perceptually challenging cases, while for the chair-full and pose tasks the models fail on easier-looking scenes. In every task, a single CMA-ES search of $10^6$ queries ($10^5$ for SAM 3D Body) finds errors far above each model's typical error; for example, SAM 3D Body's worst case has a joint error of over 1\,m. For both chair tasks, the worst-case error is greater than $270\times$ the median error. Table~\ref{tab:sota} in the appendix presents the numeric results. Thus, our tasks are sufficiently challenging to expose severe failures of SOTA models.

\subsection{Training Sampling Concentration}

 \begin{figure}
 \centering
 \includegraphics[width=0.7\linewidth]{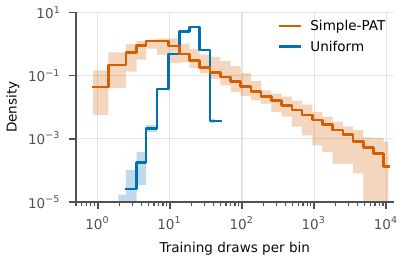}
 \caption{Distribution of training draws per bin. The highlights display the min-max range over 5 seeds and 8 coordinate subsets for each seed.}
 \label{fig:overconc_hist}
 \vspace{-1em}
 \end{figure}
To quantify how strongly Simple-PAT concentrates samples in particular regions of the parameter space, we examine the distribution of samples from a training run for the normal estimation task. We partition the parameter space into $8^6 = 262{,}144$ bins by selecting six coordinates and dividing each $[0,1]$ interval into 8 equal-length sections. 

Figure \ref{fig:overconc_hist} shows the resulting distribution of counts. Under uniform sampling, the distribution is tightly concentrated around its mode of 19 samples per bin, consistent with a Poisson distribution. Under adversarial sampling, the mode is only 5 samples per bin, but with a long tail extending over several orders of magnitude. This concentration is consistent with these cases being particularly challenging for the network to learn and therefore requiring many more samples.

Compared to uniform sampling, Simple-PAT over-samples the top 1\% and 0.1\% of bins by factors of $40\times$ and $176\times$, respectively. Therefore uniform training would need to extend $40\times$ or $176\times$ longer to reach equal exposure to these challenging cases.

\section{Conclusion}
We study procedural adversarial learning (PAL), in which a model is trained against an adversary that searches the parameters of a procedural generator, and is then judged by the worst case a strong adversary can find in that same space. We design a Simple-PAT algorithm that realizes this adversarial game with zeroth-order CMA-ES optimization on the parameter space. Across our four PAL-Dojo tasks, Simple-PAT improves the robustness over uniform training and robust reweighting variants. However, Simple-PAT exhibits some training instabilities and can miss failure modes. Future work might look to improve performance and training stability or assess how robustness in synthetic settings transfers to the real world.

\section{Acknowledgments} This work was partially supported by the National Science Foundation.

{
    \small
    \bibliographystyle{ieeenat_fullname}
    \bibliography{main}
}

\newpage

\appendix

\section{Hyperparameters}
\label{sec:app_hparam}

\begin{itemize}
    \item \textbf{Training:} each model is trained for 20,000 steps at batch size 256.
    \item \textbf{Models:} for Normals and chair tasks we train small U-nets with 7.5M and 7.8M parameters, respectively. These networks consist of standard blocks with two $3\times 3$ convolutions followed by a GroupNorm and ReLU. We select GroupNorm instead of BatchNorm since our adaptive sampling procedure may affect the batch statistics. The pose network has a similar structure but with only 5M parameters, since the upsampling layers are replaced with a global average pooling and output head. 
    \item \textbf{Optimizers:} we use Adam with a learning rate of $3\times 10^{-4}$ or Muon with a learning rate of 0.02. Muon is applied to the matrix and convolution parameters except the output head, and Adam is applied to the remaining parameters. Every training method uses Muon besides "Uniform (Adam)." We use a cosine learning rate schedule with a warmup of 200 steps.
    \item \textbf{Simple-PAT Training:} We use $k = 16, \lambda = 14,$ and $n_0 = 32$. Hence, each batch is composed of 32 uniform samples and 14 samples from 16 CMA-ES instances. Every $R=40$ training steps we replace the least fit 25\% of instances. 

    The replacement is determined by the fitness function $f = f_{\mathrm{error}} + f_\mathrm{separation}$. Both terms are normalized between 0 and 1. For the error term, we use the worst error from $\pi_i$'s last generation, scaled so that the mean error on the uniform samples is 0 and the overall worst found error is 1. For the separation term, we use the distance between $\mu_i$ and its nearest neighbor $\mu_j$. We scale this by $\sqrt{d/6}$, the RMS distance between two uniform random points in $[0,1]^d$, and we clamp it between 0 and 1. Thus, the most fit $\pi_i$ samplers are those in regions with high error and far from others.
    \item \textbf{CVaR Training:} for all tasks we use $\alpha = 0.05$, so that 13 scenes contribute to the loss used to train the network.
    \item \textbf{Tilted Training:} for each task we use a converged uniform-Muon model to compute the ESS 
    ($\mathrm{ESS} = 1 / \sum_i w_i^2$, where $w_i = e^{\beta \ell_i} / \sum_j e^{\beta \ell_j}$ is the weight of scene $i$ in a batch).
    We select $\beta$ to match CVaR's ESS of 13. We compute the average over 64 batches, finding $\beta = 2490$ (Normals), 18.8 (pose), 10600 (chair-full), and 21600 (chair-simple).
\end{itemize}

\subsection{Evaluation Search}
The evaluation is composed of six searches which each use $10^6$ queries to try to maximize the model's evaluation loss over the generator's parameter space $[0,1]^d$. For each model, we take the maximum over each of the searches. Then we perform the cross-model evaluation where each model is additionally tested on the worst cases found against every other model.

We use the following searches:
\begin{itemize}
\item \textbf{CMA-ES}~\citep{cmaes}: an evolution strategy that samples each generation from a multivariate Gaussian and adapts the mean and variance towards the best samples. We use population 512, a full covariance matrix, and IPOP restarts.
\item \textbf{DIRECT-L}~\citep{directL}: a deterministic search that recursively divides the search space into rectangular regions and samples the center. We use the locally biased DIRECT-L variant.
\item \textbf{TuRBO}~\citep{turbo}: Bayesian optimization that fits a Gaussian process to previous queries and proposes new ones within a trust region around the best point; it uses restarts when the trust region collapses. We use batches of 256 points and fit the Gaussian process to at most 1024 points from the current restart.
\item \textbf{Differential evolution}~\citep{storn1997differential}: an evolutionary algorithm that proposes candidates by adding the scaled difference of two population members to a third. We use SciPy's \texttt{best1bin} strategy with population 500 for 2,000 generations. 
\item \textbf{Genetic algorithm}~\citep{pymoo}: an evolutionary algorithm that forms each generation by selecting, recombining, and mutating members of the previous generation. We use population 512 with \texttt{pymoo}'s defaults of binary tournament selection, simulated binary crossover, and polynomial mutation.
\item \textbf{Uniform sampling}: $10^6$ uniform random samples from $[0,1]^d$. This acts as a baseline to ensure that searches are working properly.
\end{itemize}

\section{Evaluation Optimizer Analysis}
\label{sec:app_optimizers}

\subsection{Model Robustness Compared to Adversary Strength}
\label{sec:app_robustness_strength}

\begin{figure*}
    \centering
    \includegraphics[width=0.9\linewidth]{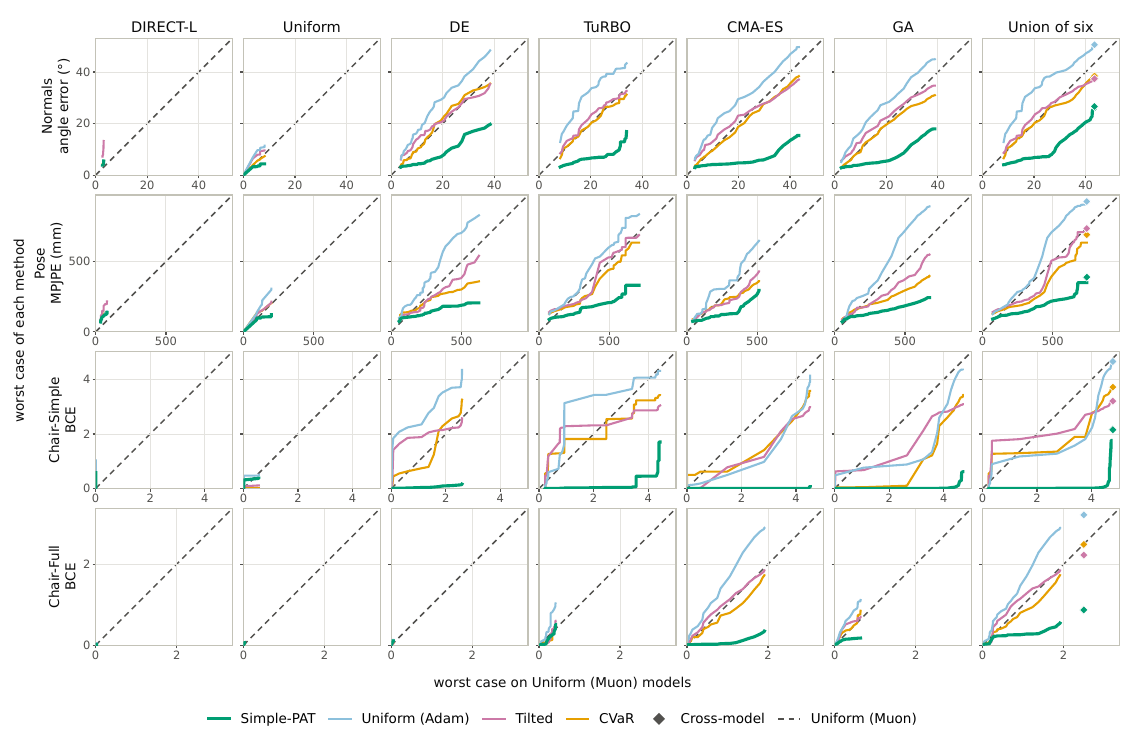}
    \caption{Worst case against each training method, with strength measured as the worst case against uniform Muon. Each line represents a single search over budgets from $10^3$ to $10^6$ queries and shows the mean worst case over all five seeds.}
    \label{fig:app_adversary_strength}
\end{figure*}

Figure \ref{fig:app_adversary_strength} describes each method's robustness by comparing their performance against attacks of various strengths, where strength is determined by the attack effectiveness against uniform-Muon. This is similar to Figure \ref{fig:adversary_strength}, except each line represents a single adversary with varying search budgets from $10^3$ to $10^6$. Specifically, for a search worst cases at step $t$, $W_t$, the line plotted is $(W_t(\textrm{Muon}), W_t(\textrm{method)})$ for $t \in [10^3,10^6]$; here we average $W_t$ over the five seeds.

In most cases, we see that searches are fairly effective against uniform-Muon (the $x$ axis reaches large values), but these searches are less effective against Simple-PAT than they are against alternate methods. With DIRECT-L and Uniform, we see instances in which the search is ineffective against all methods. Another interesting case is the Chair-Full TuRBO search, which appears similarly effective against Simple-PAT as alternate methods. However, even against Simple-PAT, TuRBO is considerably less effective than the cross-model search (0.51 BCE compared to 0.87 BCE).

\subsection{Efficacy of Different Optimizers}

\begin{table}[h]
\centering\small
\begin{tabular}{@{}l r r r r@{}}
\toprule
Task & CMA-ES & DE & TuRBO & GA \\
\midrule
Normals      & 17 & 7  & 0  & 1\\
Chair-Simple & 4  & 9  & 5  & 7\\
Chair-Full   & 24 & 0  & 1  & 0\\
Pose         & 0  & 7  & 11 & 7\\
\midrule
Total        & 45 & 23 & 17 & 15\\
\bottomrule
\end{tabular}
\caption{Search that found each cross-model worst case (25 models per task). DIRECT-L and uniform sampling found none.}
\label{tab:source_counts}
\end{table}

Table~\ref{tab:source_counts} displays which optimizer found the cross-model worst case for each task and training method. CMA-ES is the most effective overall, though DE, TuRBO, and GA also find worst cases between 15\% and 23\% of the time. See Table \ref{tab:per_optimizer} for a more detailed analysis of the worst cases found by each search against each training method. On a practical note, we find that the DIRECT-L and TuRBO methods are most expensive. For a shorter evaluation (e.g. for algorithm development) we recommend running only the CMA-ES, DE, and GA algorithms. This produces the same significance results on every task, and 87 of the 100 models have a cross-model worst case within 5\% of the six-search value. However, it misses the TuRBO searches that find Simple-PAT's largest failures on Chair-Simple.

\begin{figure}[H]
    \centering
    \includegraphics[width=0.9\linewidth]{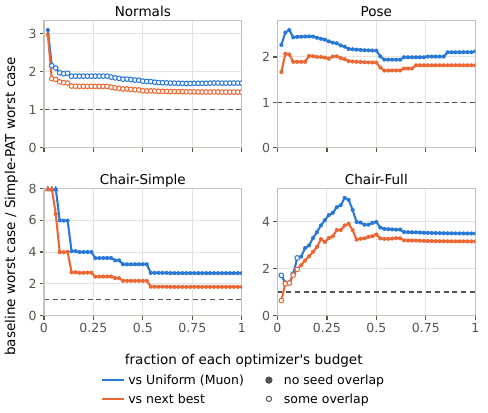}
    \caption{Ratio of the baseline evaluation worst case compared to the Simple-PAT evaluation worst case as the
     search budget increases. At each budget fraction $f$, each model's worst case is the maximum over all six
     searches each at fraction $f$ of their total budget of 1M samples. ``No seed overlap'' indicates that every Simple-PAT seed is better than every seed of the baseline.}    
    \label{fig:ratio_vs_budget}
\end{figure}

As discussed in Section \ref{sec:eval_practical}, if an optimizer has stalled, we cannot claim that it has converged. However, comparisons that are changing at the budget cutoff would indicate that a larger evaluation budget is likely to change conclusions. Figure \ref{fig:ratio_vs_budget} shows that this is not the case. Over the final quarter of the budget, the ratio between Simple-PAT and the next-best baseline changes by less than $1\%$ on every task. Appendix \ref{sec:app_convergence_curves} additionally shows each search's progress as a function of the number of queries.

\begin{table*}[h]
\centering
\small
\setlength{\aboverulesep}{0pt}\setlength{\belowrulesep}{0pt}
\begin{tabular}{@{}l l r r r r r@{}}
\toprule
Task & Search & Adam & Muon & Tilted & CVaR & Simple-PAT \\
\midrule
Normals & Uniform ($10^6$) & 11.4 & 8.22 & 9.54 & 7.35 & 4.39 \\
angle ($^\circ$) & DIRECT-L & 4.14 & 3.13 & 13.2 & 4.26 & 5.47 \\
 & Diff.\ evolution & 48.4 & 38.3 & 35.3 & 35.5 & \textbf{19.7} \\
 & CMA-ES & \textbf{49.7} & \textbf{43.4} & \textbf{37.1} & \textbf{38.4} & 15.4 \\
 & TuRBO & 43.4 & 33.9 & 32.7 & 31.2 & 16.9 \\
 & Genetic alg. & 44.9 & 38.9 & 34.7 & 31.0 & 17.9 \\
\cmidrule(l){2-7}
 & Max (single-model) & 50.5 & 43.4 & 37.4 & 38.4 & 25.6 \\
\midrule
Chair-Simple & Uniform ($10^6$) & 0.505 & 0.567 & 0.116 & 0.023 & 0.383 \\
BCE & DIRECT-L & 1.03 & 0.010 & 0.044 & 0.648 & 0.580 \\
 & Diff.\ evolution & 4.35 & 2.59 & 2.57 & 3.26 & 0.161 \\
 & CMA-ES & 4.13 & 4.52 & 2.99 & \textbf{3.58} & 0.072 \\
 & TuRBO & 4.32 & 4.44 & 3.05 & 3.42 & \textbf{1.71} \\
 & Genetic alg. & \textbf{4.36} & \textbf{4.70} & \textbf{3.10} & 3.43 & 0.619 \\
\cmidrule(l){2-7}
 & Max (single-model) & 4.51 & 4.71 & 3.19 & 3.62 & 1.77 \\
\midrule
Chair-Full & Uniform ($10^6$) & 0.064 & 0.028 & 0.026 & 0.025 & 0.064 \\
BCE & DIRECT-L & 0.050 & 0.025 & 0.022 & 0.017 & 0.019 \\
 & Diff.\ evolution & 0.130 & 0.050 & 0.126 & 0.051 & 0.107 \\
 & CMA-ES & \textbf{2.90} & \textbf{1.91} & \textbf{1.85} & \textbf{1.73} & 0.349 \\
 & TuRBO & 1.03 & 0.407 & 0.602 & 0.468 & \textbf{0.511} \\
 & Genetic alg. & 1.12 & 0.638 & 0.700 & 0.853 & 0.177 \\
\cmidrule(l){2-7}
 & Max (single-model) & 2.90 & 1.91 & 1.85 & 1.73 & 0.547 \\
\midrule
Pose & Uniform ($10^6$) & 307 & 198 & 215 & 205 & 122 \\
MPJPE (mm) & DIRECT-L & 146 & 80.8 & 217 & 138 & 139 \\
 & Diff.\ evolution & 829 & 624 & 541 & 359 & 206 \\
 & CMA-ES & 647 & 513 & 429 & 360 & 294 \\
 & TuRBO & 836 & \textbf{711} & \textbf{688} & \textbf{634} & \textbf{331} \\
 & Genetic alg. & \textbf{893} & 674 & 550 & 398 & 243 \\
\cmidrule(l){2-7}
 & Max (single-model) & 920 & 742 & 727 & 635 & 350 \\
\bottomrule
\end{tabular}

\caption{Single-model worst case found by each evaluation search, mean over 5 seeds. Bold marks the strongest individual search for each method.}
\label{tab:per_optimizer}
\end{table*}

\section{Convergence of the Evaluation Search}
\label{sec:app_convergence}
Let $W_n$ represent the worst case found by the evaluation search after $n$ samples. This section details the following claims (which hold under the assumptions below): (1) as $n$ approaches infinity, $W_n$ approaches the model's true worst case on $[0,1]^d$, (2) Let $M$ be the largest possible error. For arbitrary $T$ and $\delta < M$ there can exist a model for which $W_{n+1} - W_n > \delta$ for some $n \geq T$.

\noindent
\textbf{Setting.} We use a fixed model with parameters $\theta$. For a generator parameter vector $u \in [0,1]^d$ let $\ell(u)$ denote the model's loss on the scene specified by $u$. The true worst case is $W = \sup_{u\in[0,1]^d} \ell(u)$. A search queries points $u_1, u_2,...$; after $n$ queries the maximum loss found is $W_n = \max_{i \leq n} \ell(u_i)$. By definition $W_n$ is increasing and $W_n \leq W$.

We can further decompose $\ell(u)$ as $\ell(u) = \textrm{error}(f_\theta(x(u)), y(u))$ where $\textrm{gen}(u) = (x(u), y(u))$.

\noindent
\textbf{Assumption 1:} $f_\theta$ and $\textrm{error}$ are continuous. 

\noindent
\textit{Justification:} up to floating point precision, the neural network components and loss functions we use are continuous.

\noindent
\textbf{Assumption 2:} $\textrm{gen}$ is piecewise continuous. There are finitely many closed regions $\mathcal{R} = \{A_1,\dots, A_m\}$ covering $[0,1]^d$, each equal to the closure of its interior, and continuous functions $g_1,\dots,g_m$ such that $\textrm{gen} = g_j$ on the interior of $A_j$, and every $u$ satisfies $\textrm{gen}(u) = g_j(u)$ for some region $A_j$ containing $u$.

\noindent
\textit{Justification:} In our procedural generators, there are two sources of discontinuity: discrete procedural choices and rendering discontinuities. Each generator has a finite number of discrete choices which are selected by comparisons with coordinates of $u$ or a continuous function of $u$ (e.g. make choice ``A'' if $u_2 < 0.3$). Therefore each discrete choice is represented in a set with nonzero volume. 

We expect that discontinuities due to rendering will not have a major effect and in practice can be treated as continuous. For the chair and pose tasks, we use the differentiable renderer nvdiffrast~\citep{nvdiffrast}. We also assume the loss varies continuously as the label boundaries move (the chair mask and joint visibility), as it would if these labels were antialiased like the image.

\noindent
\textbf{Lemma 1:} \emph{$\ell$ is piecewise continuous on the same regions $\{A_1,\dots, A_m\}$.} 

\noindent\textbf{Proof:}
Let $h_j = \textrm{error}(f_\theta(g_j))$, which is continuous by Assumption~1. By Assumption~2, $\ell = h_j$ on the interior of $A_j$, and every $u$ satisfies $\ell(u) = h_j(u)$ for some region $A_j$ containing $u$.

\noindent
\textbf{Lemma 2:} \emph{For every $\epsilon > 0$, the set of scenes with $\ell > W - \epsilon$ contains a nonempty open set $O_\epsilon$.} 

\noindent\textbf{Proof:}
Since each $h_j$ is continuous on the compact set $A_j$ there is a region $A_j$ and a point $u^* \in A_j$ with $h_j(u^*) = W$. By continuity there is an open set $U \ni u^*$ with $h_j > W - \epsilon$ on $U$. Let $O_\epsilon = U \cap \mathrm{int}(A_j)$. It is nonempty since $A_j$ is the closure of its interior.

\noindent
\textbf{Lemma 3:} \emph{Each of the following searches eventually queries any nonempty open set $O$ (with probability one for randomized searches).}

\noindent\textbf{Proof:}
\begin{itemize}
    \item \emph{Uniform sampling}: the probability of missing $O$ is less than one at each step, and the probability of missing at all steps goes to zero as $n \to \infty$.
    \item \emph{DIRECT-L}: one of the largest rectangles is divided at every iteration, so all rectangles shrink to zero size. Eventually a rectangle lies inside $O$ and its center is queried.
    \item \emph{TuRBO}: TuRBO will restart infinitely often -- improvements must exceed a fixed fraction of a bounded loss, so they will eventually stop and the trust region will shrink until it restarts. At each restart, it selects new points using a Sobol sequence over the whole cube, which will eventually hit every open set.
    \item \emph{Full evaluation:} it reports the maximum over multiple searches, including the above.
\end{itemize}

\noindent
\textbf{Claim 1:} \emph{Under Assumptions 1 and 2, $W_n \to W$ as $n \to \infty$ for each search in Lemma~3 (with probability one for randomized searches).} 

\noindent\textbf{Proof:}
Fix $\epsilon > 0$. By Lemma~2 there is a nonempty open set $O_\epsilon$ on which $\ell > W - \epsilon$, and by Lemma~3 the search queries some $u_k \in O_\epsilon$. Hence $W - \epsilon < W_n \le W$ for all $n \ge k$.

\noindent \textbf{Claim 2:} \emph{For arbitrary $T$ and $\delta < M$, there can exist a model satisfying Assumptions 1 and 2 for which $W_{n+1} - W_{n} > \delta$ for some $n \geq T$.}

\noindent\textbf{Proof:}
We construct such a model -- it has a loss of zero everywhere except some closed region $A$ where it has loss greater than $\delta$. Consider the search that evaluates a model which has loss zero everywhere (with a fixed seed for randomized searches). It will query points $u_1, \dots, u_T$. We choose a set $A$ (closed with nonempty interior) such that $u_i \not\in A$ for all $i$. From Lemma 3, the search will eventually query some point in $A$; suppose this first occurs at step $n+1$, then we have $W_{n+1}-W_n > \delta$.

\section{Evaluation with a Single-Model Adversary}
\label{sec:app_single}

\begin{table*}[h]
\centering
\small
\setlength{\aboverulesep}{0pt}\setlength{\belowrulesep}{0pt}
\renewcommand{\arraystretch}{1.18}
\setlength{\tabcolsep}{3.5pt}
\begin{tabular}{@{}l | r@{\,}l | r@{\,}l | r@{\,}l | r@{\,}l@{}}
\toprule
Method & \multicolumn{2}{c|}{Normals, angle ($^\circ$)} & \multicolumn{2}{c|}{Chair-Simple, BCE} & \multicolumn{2}{c|}{Chair-Full, BCE} & \multicolumn{2}{c}{Pose, MPJPE (mm)} \\
\midrule
Uniform (Adam) & 50.55 & {\scriptsize$\pm$0.70}$^*$ & 4.51 & {\scriptsize$\pm$0.15}$^*$ & 2.90 & {\scriptsize$\pm$0.81}$^*$ & 920 & {\scriptsize$\pm$51}$^*$ \\
Uniform (Muon) & 43.4 & {\scriptsize$\pm$3.0} & 4.712 & {\scriptsize$\pm$0.076}$^*$ & 1.91 & {\scriptsize$\pm$0.66}$^*$ & 742 & {\scriptsize$\pm$33}$^*$ \\
Tilted & 37.4 & {\scriptsize$\pm$1.5} & 3.19 & {\scriptsize$\pm$0.14}$^*$ & 1.85 & {\scriptsize$\pm$0.24}$^*$ & 730 & {\scriptsize$\pm$100}$^*$ \\
CVaR & 38.4 & {\scriptsize$\pm$2.2} & 3.62 & {\scriptsize$\pm$0.23}$^*$ & 1.73 & {\scriptsize$\pm$0.83}$^*$ & 640 & {\scriptsize$\pm$130}$^*$ \\
\textbf{Simple-PAT} & \textbf{26} & {\scriptsize$\pm$11} & \textbf{1.77} & {\scriptsize$\pm$0.60} & \textbf{0.55} & {\scriptsize$\pm$0.17} & \textbf{350} & {\scriptsize$\pm$41} \\
\bottomrule
\end{tabular}

\caption{Worst case found by the single-model evaluation search ($6\times10^6$ queries per model): mean $\pm$ 1 standard deviation over 5 seeds. Bold marks the best method in each column. $^*$: Simple-PAT's worst case is significantly lower than this baseline's (two-sided Mann-Whitney test, 5 vs.\ 5 seeds, $p<0.05$; every marked comparison has $p=0.0079$).}
\label{tab:single_model}
\end{table*}

In Table \ref{tab:single_model} we show the complete single-model evaluation results. These largely follow the trend of the cross-model evaluation. Under single-model evaluation, Simple-PAT's advantage over the strongest baseline is larger on chair-full ($3.16\times$ vs.\ $2.55\times$ cross-model) and chair-simple ($1.81\times$ vs $1.49\times$ cross-model); on pose and normals Simple-PAT's advantage over the strongest baseline is similar when using the single-model or cross-model evaluation. Thirteen of the sixteen comparisons are significant at $p = 0.05$, each with $p = 0.0079$, meaning that every Simple-PAT seed's worst case is better than every seed of that baseline. The exceptions are normals against Muon, tilted and CVaR ($p = 0.056$, $0.15$ and $0.15$), where the seeds overlap.

\section{Procedural Generator Design}
\label{sec:app_gen_design}

When designing procedural generators, the searchability of the space impacts both the efficacy of the Simple-PAT algorithm and the evaluation. Here we present results using prototype versions of the environments described in the paper. These environments differ from the versions described in Section \ref{sec:procedural_generators} in the following ways:

\begin{itemize}
   \item \textbf{Object and room color.} In the prototype chair environments, each object's color was a hash of the slot's other parameters (position, scale, appearance, yaw and tilt), and the floor and wall colors were a hash of the room dimensions and camera azimuth. Any change to those parameters therefore redrew the colors in a seemingly random pattern. In the final environments, object color depends only on that object's appearance parameter, and room appearance is a single parameter.
   \item \textbf{Object class.} The prototype chair environments selected objects through shared composition and permutation parameters, so one parameter could exchange the objects in several slots at once. The final environments give each slot its own class parameter.
   \item \textbf{Body pose.} The prototype pose environment chose a stance with a discrete selector parameter and drew the body from a separate PCA model for each stance. The final environment uses a single archetypal model that allows continuous interpolation between all poses.
   \item \textbf{Dimension.} Chair-simple grows from 11 to 13 dimensions, chair-full from 54 to 69
   and pose from 44 to 49.
\end{itemize}
Training, search, and evaluation of the prototype environments follow the same protocol described in the main paper with very minor implementation differences\footnote{In the prototype experiments, the pose loss computed the root-relative joint positions in \texttt{bf16} instead of \texttt{fp32}. In the prototype runs, a Simple-PAT search instance was replaced with an instance initialized from the farthest of 32 uniform draws from the other members' means; in the final runs it is initialized from a uniformly random point. This change was made for simplicity.}.

\begin{figure}[t]
   \centering
   \includegraphics[width=0.9\linewidth]{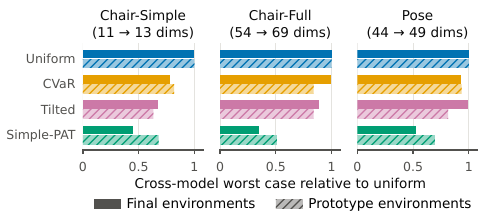}
   \caption{Cross-model worst case of each method relative to uniform-Muon training, in the prototype (hatched) and final (solid) environments. Each bar is the mean over five seeds divided by the mean for uniform training in the same version.}
   \label{fig:prototype_ratio}
 \end{figure}

We first consider the efficacy of Simple-PAT. Figure \ref{fig:prototype_ratio} shows the reduction in cross-model worst-case error by the Simple-PAT algorithm in both versions. In the prototype environments, Simple-PAT does not outperform tilted training on chair-simple, and it achieves significantly better results at $p=0.05$ on six of the twelve comparisons on the three changed tasks. The comparisons not found to be significant are tilted loss and CVaR on chair-simple and all four comparisons on pose. On the final generators, Simple-PAT is found to be better than every baseline on these three tasks, with $p=0.0079$ for all twelve comparisons. Thus, improving the searchability of the space improves Simple-PAT's advantage in this case. 

\begin{figure}
    \centering
    \includegraphics[width=0.9\linewidth]{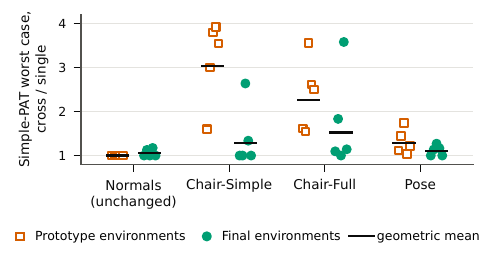}
    \caption{The ratio between the cross-model worst case and single-model worst case evaluation result between the prototype and final environments. With an effective single-model search this ratio should be close to 1.}
    \label{fig:prototype_search_gap}
\end{figure}

The generator design also affects the reliability of the evaluation. Discontinuities in the environment had the effect of hiding model failures from the evaluation search, as shown in Figure \ref{fig:prototype_search_gap}. In the prototype chair-simple environment, the Simple-PAT single-model evaluation found a mean worst case of 1.13 BCE, but the cross-model search found 3.11 average BCE on the same models. In the final environment, the single-model average worst case was 1.77, compared to 2.16 cross-model evaluation average. Among the final environments, chair-full displays the largest discrepancy between the single-model and cross-model evaluations, which is consistent with the difficulty of searching in 69 dimensions.

\section{Statistical Analysis}
\label{sec:app_stats}

Here, we discuss the power of the original five-seed test used in Section \ref{sec:cross_model_eval}. We specifically analyze the Normals environment where we observe considerable noise between seeds. We claim that if the baselines' worst cases were $1.5\times$ that of Simple-PAT, then, given the noise observed in the five original seeds, our Mann-Whitney U test would have a power of 0.33. This may explain why we were unable to reject the null hypothesis in the five-seed test, while the fifteen-seed test rejects the null hypothesis against each baseline with $p \leq 0.002$.

We estimate the power of the five-seed Normals test using a simulation. For each method, we divide the five seeds' cross-model worst cases by their geometric mean. A simulated experiment draws $n$ of these normalized values with replacement for Simple-PAT and a baseline, multiplies the baseline's draws by 1.5, and applies the two-sided Mann-Whitney U test at $p < 0.05$. Over 4,000 simulated experiments, averaged over the four baselines, the test detects the difference with probability 0.33 at $n=5$, 0.69 at $n=10$, and 0.84 at $n=15$.

\section{State-of-the-Art Foundation Model Worst-case}
\label{sec:app_sota}

\begin{table}[h]
\centering
\footnotesize\setlength{\tabcolsep}{3pt}
\setlength{\aboverulesep}{0pt}\setlength{\belowrulesep}{0pt}
\renewcommand{\arraystretch}{1.18}
\begin{tabular}{@{}l l r@{\,}l | r@{\,}l@{}}
\toprule
 & & \multicolumn{2}{c|}{Search} & \multicolumn{2}{c}{Test set} \\
\cmidrule(r){3-4}\cmidrule(l){5-6}
Task & Model & \multicolumn{2}{c|}{Worst case $\downarrow$} & \multicolumn{2}{c}{Median $\downarrow$} \\
\midrule
Normals & MoGe-2 (64 px) & 82.8 & & 22.2 & \\
angle ($^\circ$) & MoGe-2 (512 px) & 69.8 & & 18.9 & \\
 & \textbf{Simple-PAT} (single) & 26 & {\scriptsize$\pm$11} & 1.713 & {\scriptsize$\pm$0.018} \\
 & \textbf{Simple-PAT} (cross) & 27 & {\scriptsize$\pm$11} & & \\
\midrule
Chair-Simple & SAM 3 & 2.98 & & 0.00172 & \\
BCE & \textbf{Simple-PAT} (single) & 1.77 & {\scriptsize$\pm$0.60} & 1.88e-5 & {\scriptsize$\pm$4.7e-6} \\
 & \textbf{Simple-PAT} (cross) & 2.16 & {\scriptsize$\pm$0.26} & & \\
\midrule
Chair-Full & SAM 3 & 0.792 & & 0.00291 & \\
BCE & \textbf{Simple-PAT} (single) & 0.55 & {\scriptsize$\pm$0.17} & 5.6e-5 & {\scriptsize$\pm$2.0e-5} \\
 & \textbf{Simple-PAT} (cross) & 0.87 & {\scriptsize$\pm$0.43} & & \\
\midrule
Pose & SAM 3D Body & 1020 & & 67.7 & \\
MPJPE (mm) & \textbf{Simple-PAT} (single) & 350 & {\scriptsize$\pm$41} & 26.9 & {\scriptsize$\pm$1.4} \\
 & \textbf{Simple-PAT} (cross) & 389 & {\scriptsize$\pm$46} & & \\
\bottomrule
\end{tabular}

\caption{Worst case evaluation of SOTA foundation models. For normals, we evaluate MoGe-2 at the generator's standard resolution of $64\times 64$, and a version of the generator rendered at $512\times 512$ to more closely match MoGe-2's standard input resolution.}
\label{tab:sota}
\end{table}

In Table \ref{tab:sota} we display the worst-case evaluation metrics for SOTA models. The worst case for SOTA models is evaluated with a single CMA-ES search with budget $10^6$; for pose we only use a budget of $10^5$ due to the computational expense of SAM 3D Body. We include Simple-PAT worst-case metrics for reference, though these values are not directly comparable since the SOTA models were not trained on our procedural generators. This demonstrates that our tasks are challenging, and the adversary can effectively find failures of SOTA models.

\clearpage
\onecolumn
\section{Single-Model Evaluation Worst-case Visualizations}
\label{sec:app_vis}
Here we show visualizations of each model's evaluation worst case found by the single-model search. We select the model with the median loss of the five seeds.

\subsection{Normals}
\begin{figure}[H]
    \centering
    \includegraphics[width=0.8\linewidth]{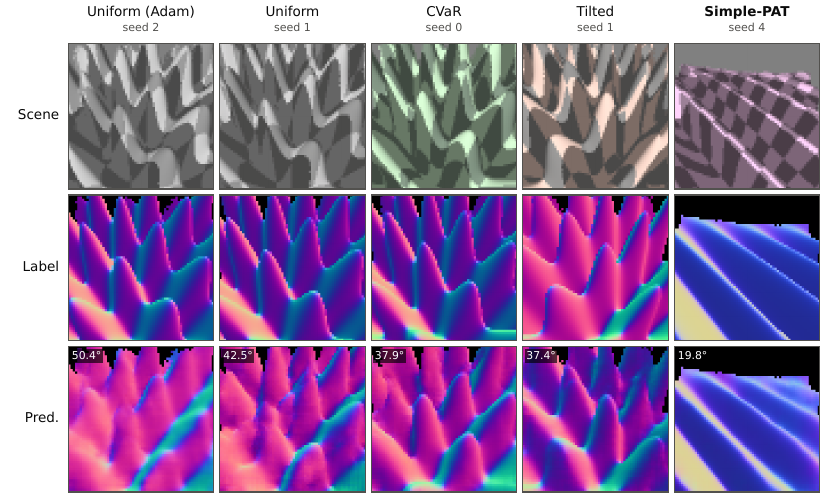}
\end{figure}

\subsection{Chair-Simple}
\begin{figure}[H]
    \centering
    \includegraphics[width=0.8\linewidth]{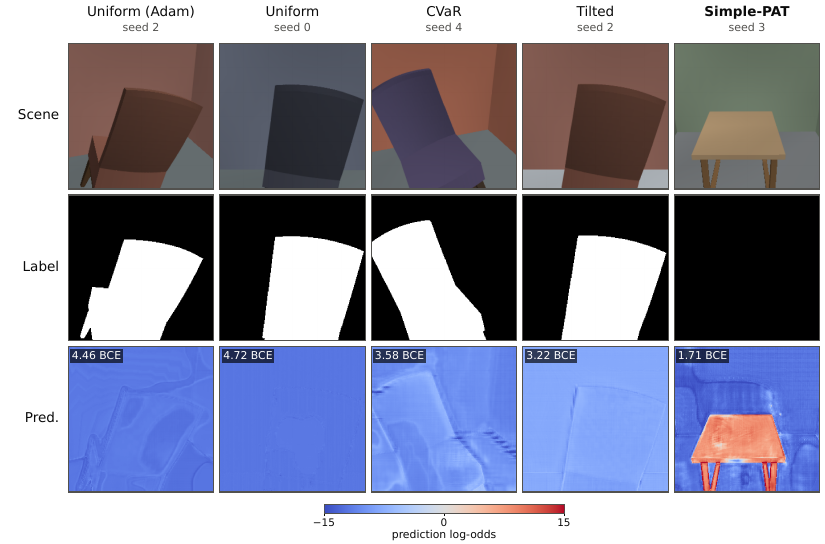}
\end{figure}

\subsection{Chair-Full}
\begin{figure}[H]
    \centering
    \includegraphics[width=0.8\linewidth]{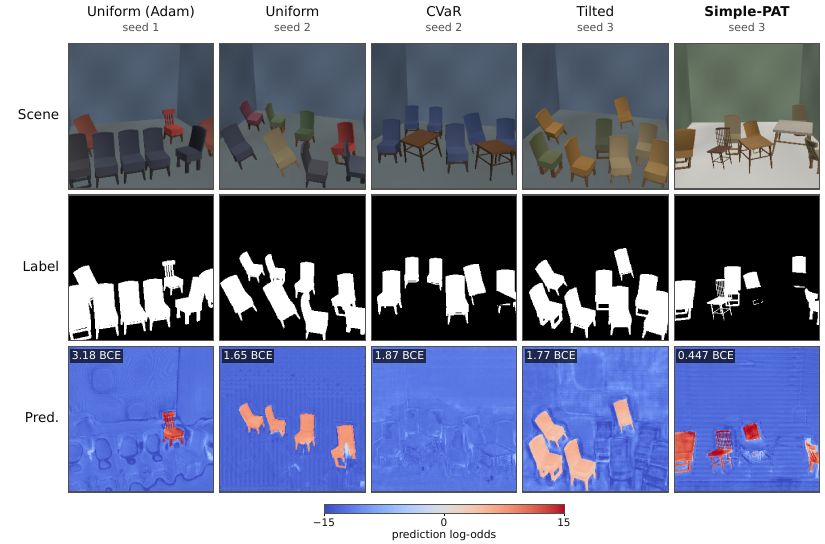}
\end{figure}

\subsection{Pose}
\begin{figure}[H]
    \centering
    \includegraphics[width=0.8\linewidth]{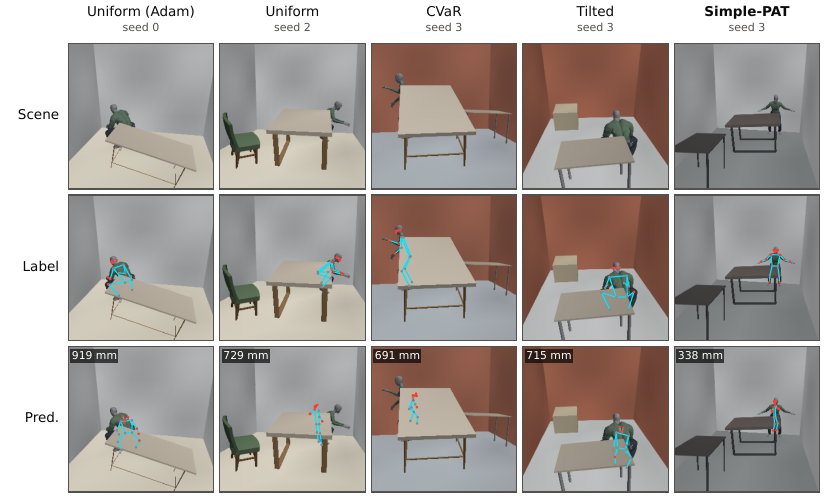}
\end{figure}

\newpage
\section{Cross-Model Evaluation Worst-case Visualizations}
Here we show visualizations of each model's evaluation worst case found by the cross-model search. We select the model with the median loss of the five seeds.

\subsection{Normals}
\begin{figure}[H]
    \centering
    \includegraphics[width=0.8\linewidth]{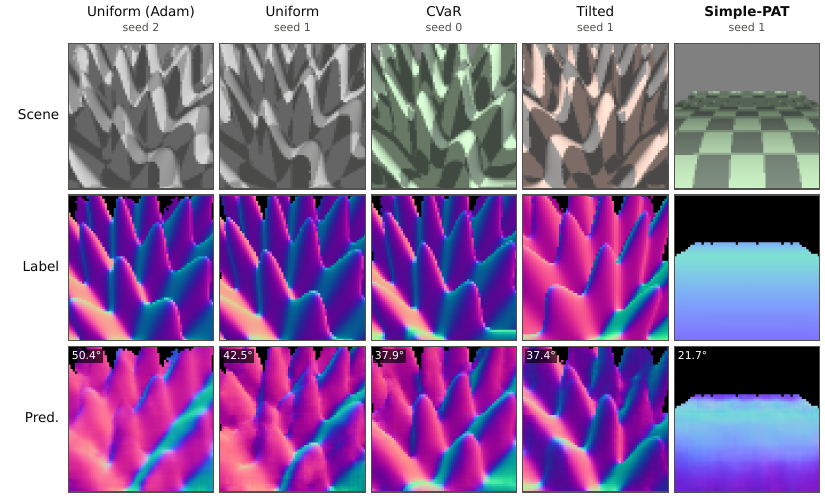}
\end{figure}

\subsection{Chair-Simple}
\begin{figure}[H]
    \centering
    \includegraphics[width=0.8\linewidth]{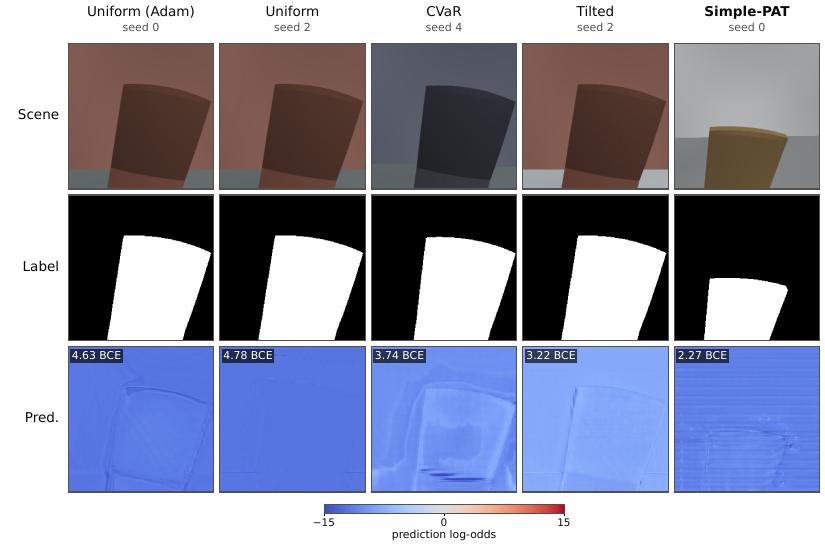}
\end{figure}

\subsection{Chair-Full}
\begin{figure}[H]
    \centering
    \includegraphics[width=0.8\linewidth]{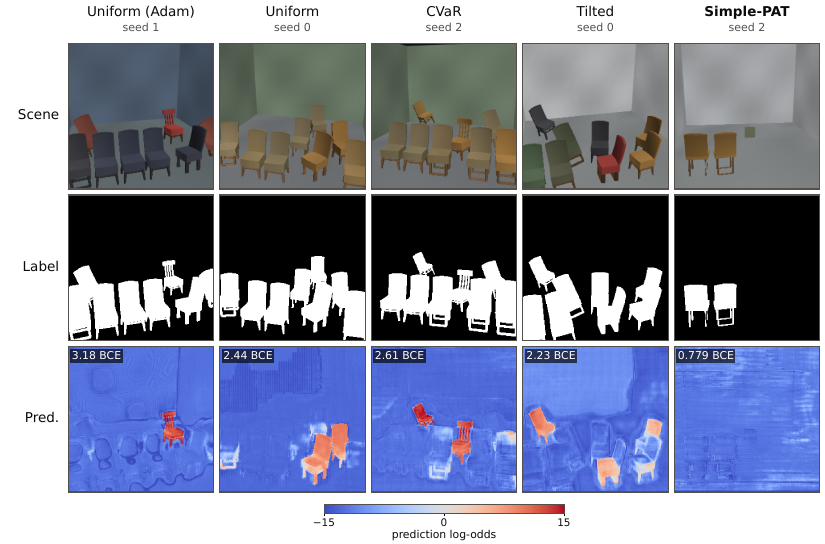}
\end{figure}

\subsection{Pose}
\begin{figure}[H]
    \centering
    \includegraphics[width=0.8\linewidth]{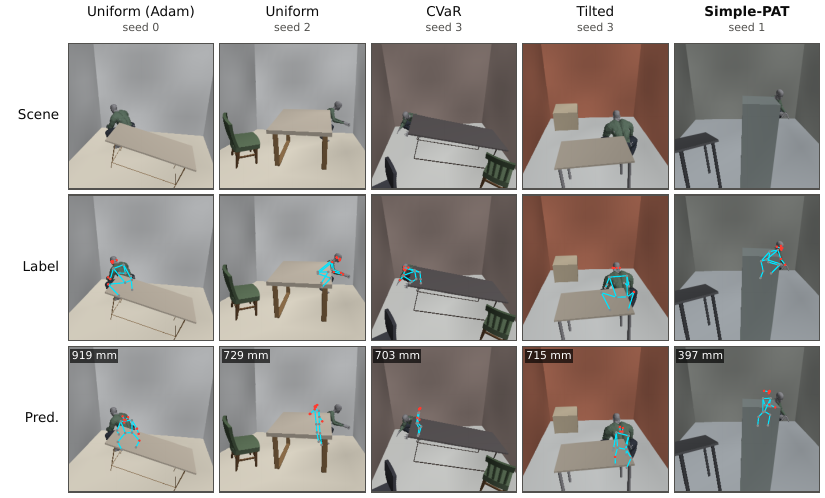}
\end{figure}

\newpage
\section{Procedural Generators}
\label{sec:app_generators}
Here we show visualizations of uniform randomly selected points from each procedural generator.

\subsection{Normals}
\begin{figure}[H]
    \centering
    \includegraphics[width=0.9\linewidth]{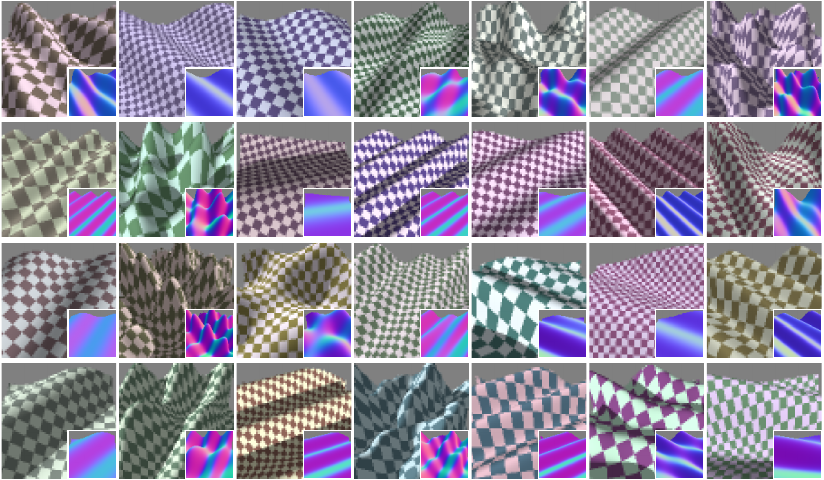}
\end{figure}

\subsection{Chair-Simple}
\begin{figure}[H]
    \centering
    \includegraphics[width=0.9\linewidth]{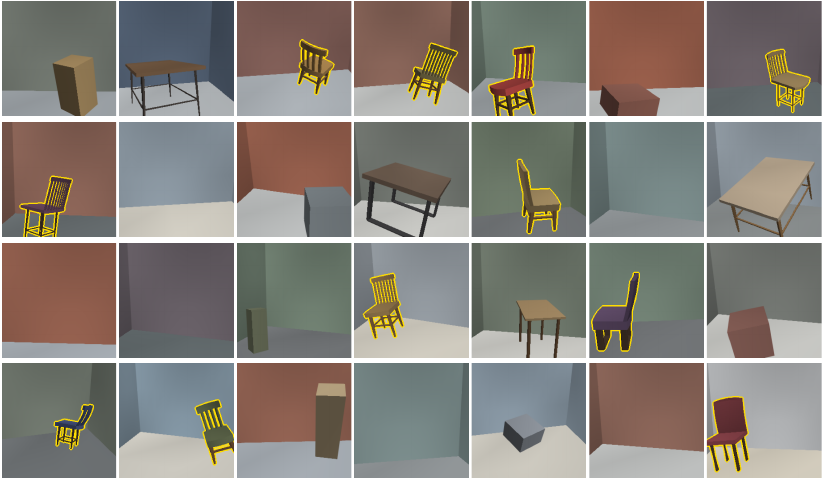}
\end{figure}

\subsection{Chair-Full}
\begin{figure}[H]
    \centering
    \includegraphics[width=0.9\linewidth]{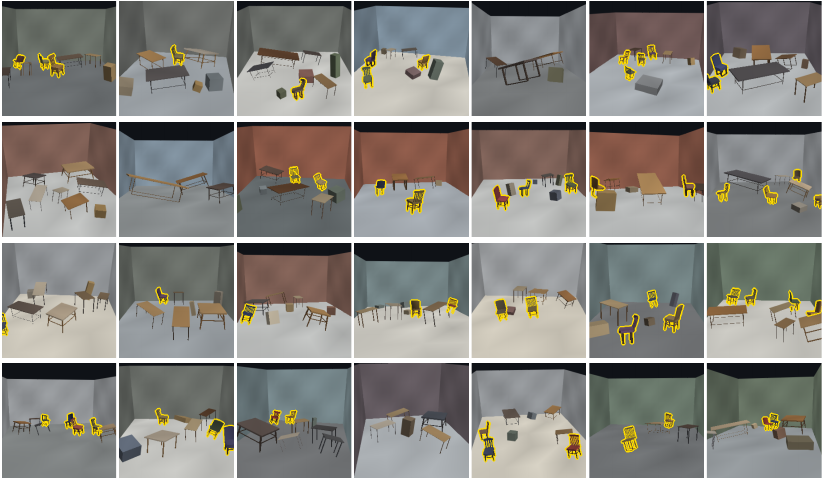}
\end{figure}

\subsection{Pose}
\begin{figure}[H]
    \centering
    \includegraphics[width=0.9\linewidth]{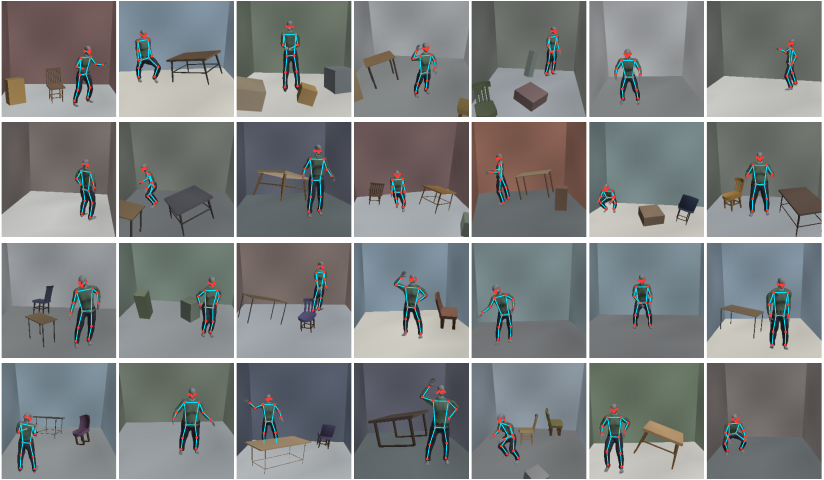}
\end{figure}

\section{Evaluation Search Progress per Trial}
\label{sec:app_convergence_curves}

Here we show the progress of each evaluation search for every trained model. Each curve represents the worst case found at each point during the search.

\subsection{Normals}
 \begin{figure}[H]
     \centering
     \includegraphics[width=\linewidth]{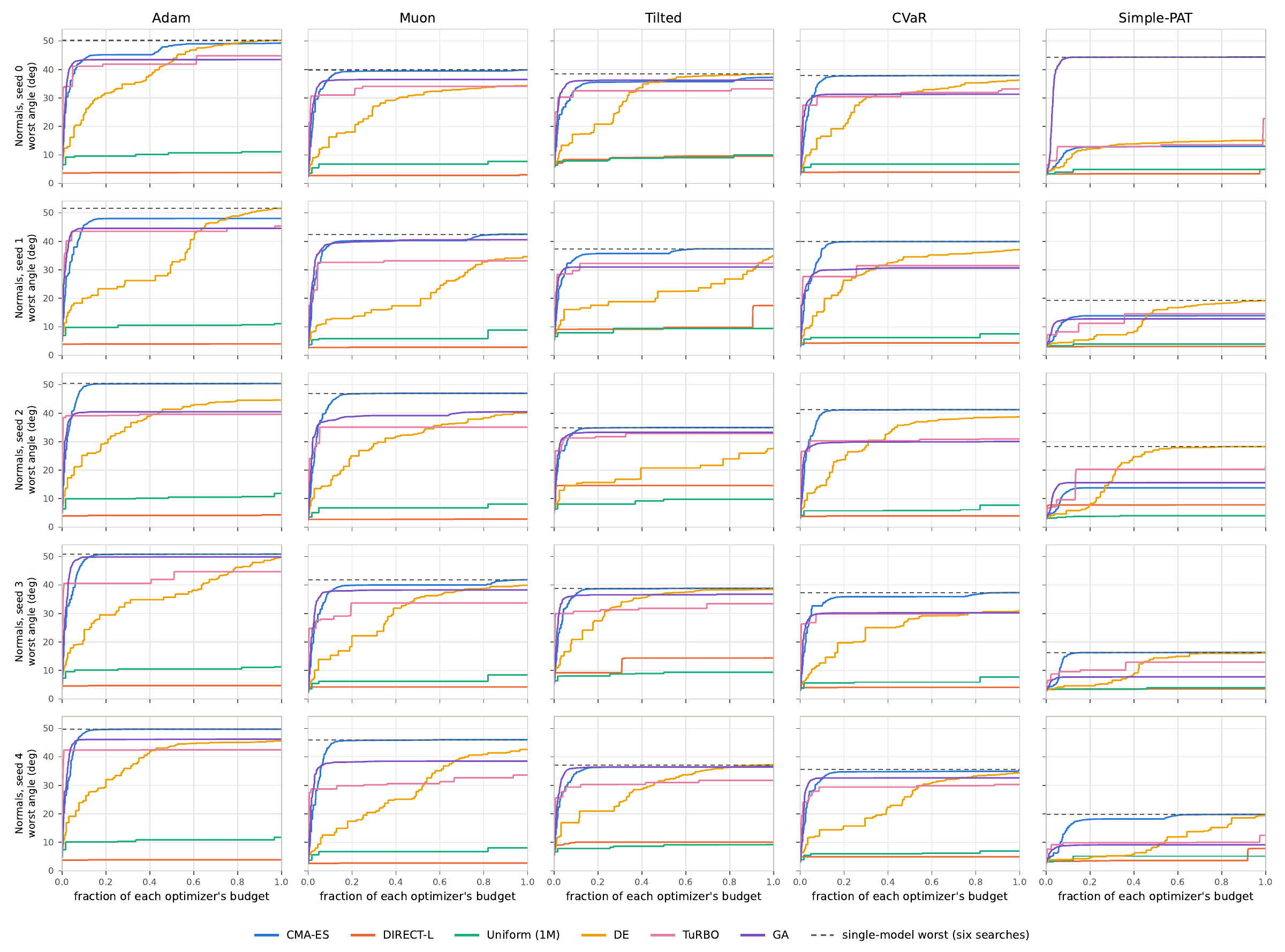}
 \end{figure}

 \clearpage
 \subsection{Chair-Simple}
 \begin{figure}[H]
     \centering
     \includegraphics[width=\linewidth]{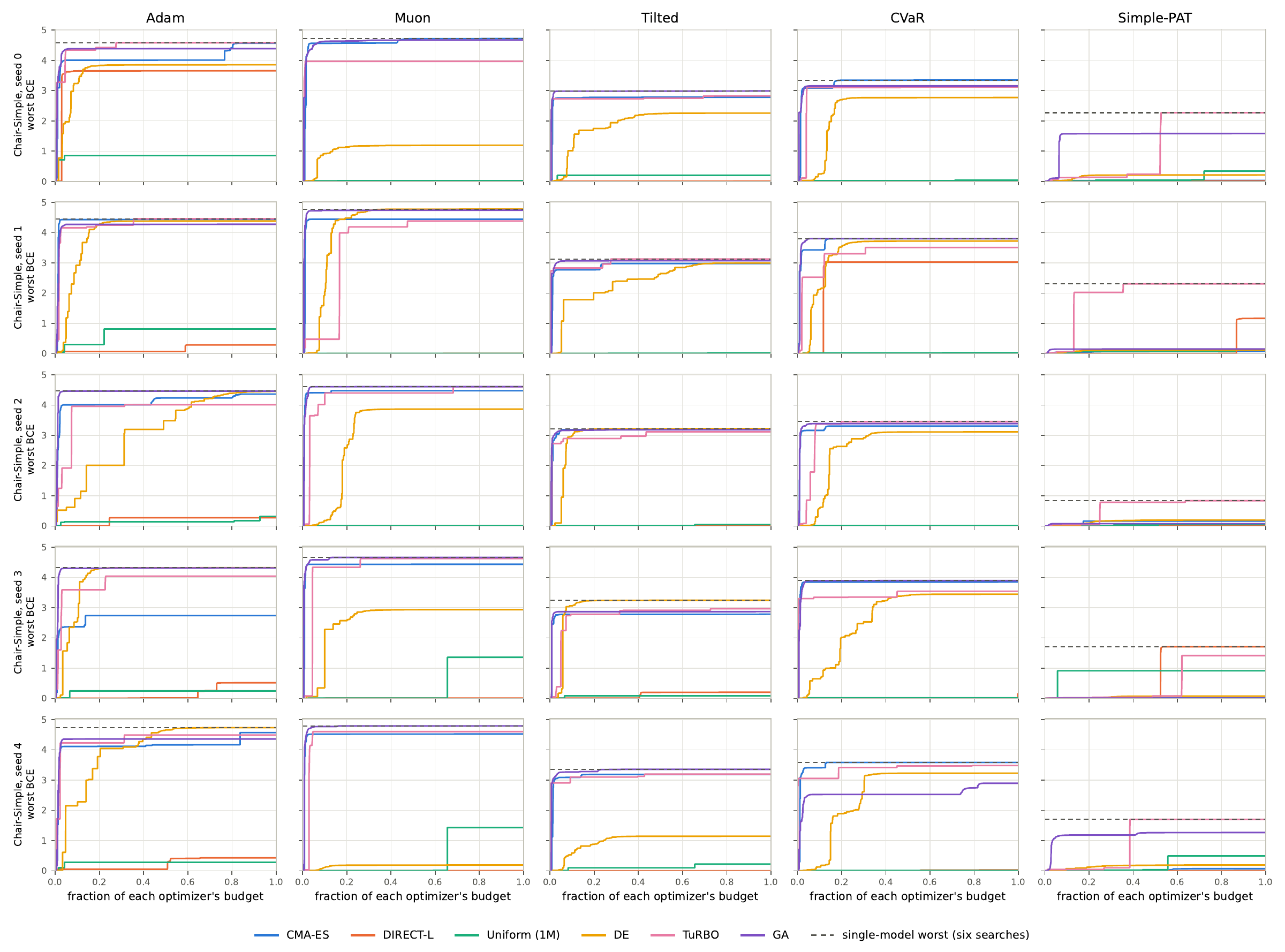}
 \end{figure}

 \clearpage
 \subsection{Chair-Full}
 \begin{figure}[H]
     \centering
     \includegraphics[width=\linewidth]{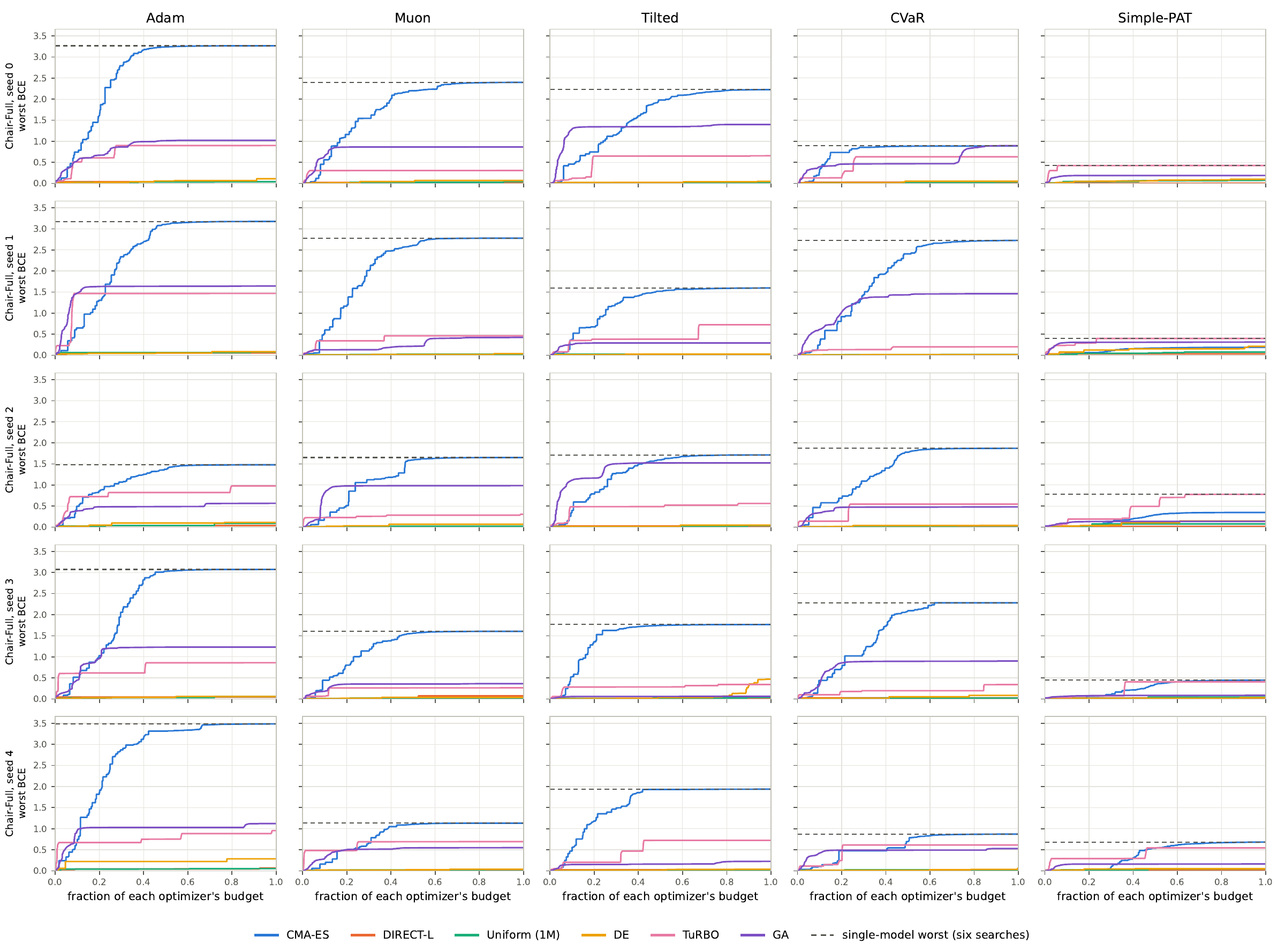}
 \end{figure}

 \clearpage
 \subsection{Pose}
 \begin{figure}[H]
     \centering
     \includegraphics[width=\linewidth]{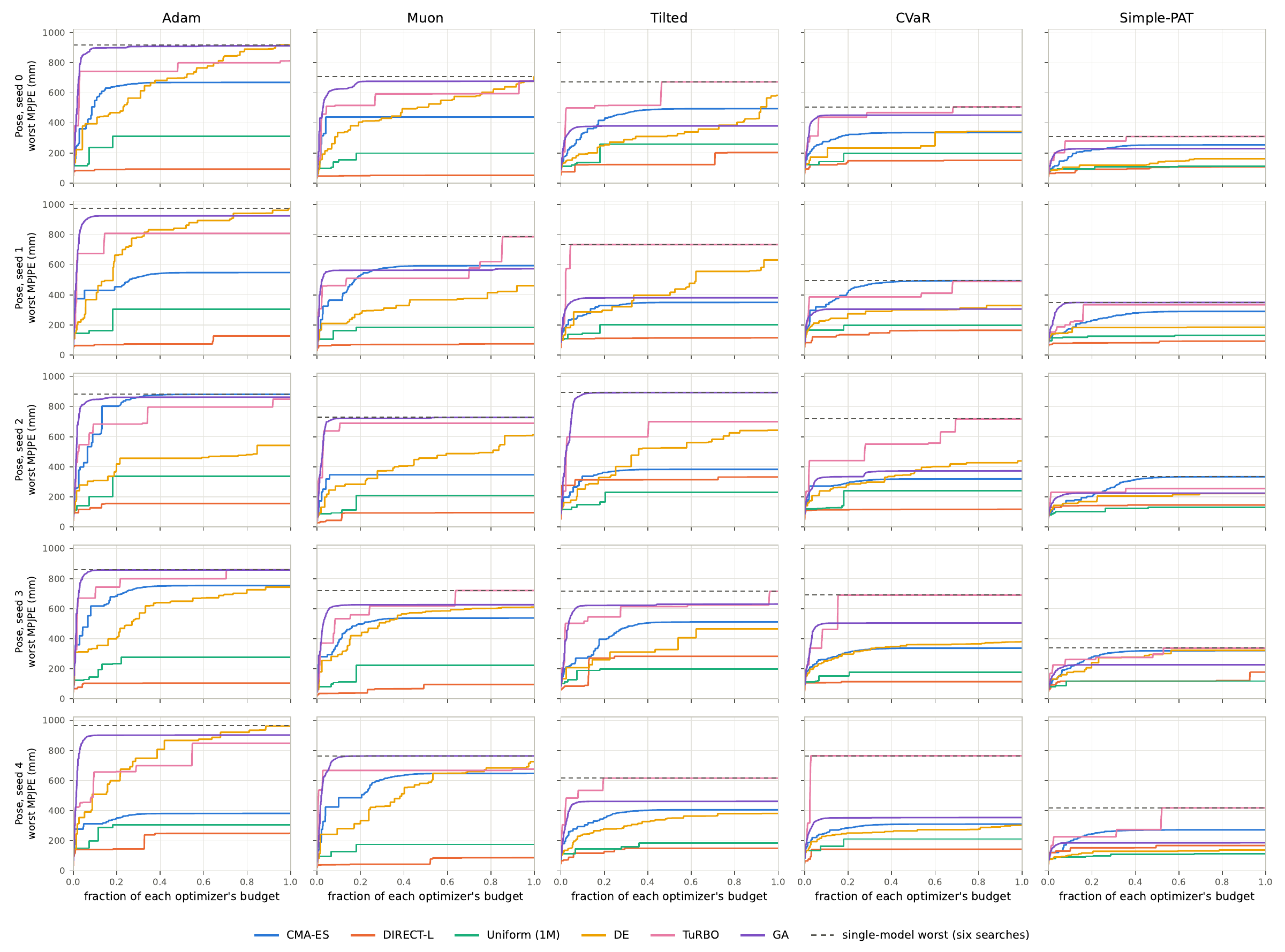}
 \end{figure}

\end{document}